\documentclass[journal,twoside,web]{ieeecolor}
\usepackage[ruled,vlined]{algorithm2e}
\usepackage{hyperref}

\usepackage{jsen}
\usepackage{cite}
\usepackage{algorithmic}
\usepackage{graphicx}
\usepackage{wrapfig}
\usepackage{multirow}

\def\BibTeX{{\rm B\kern-.05em{\sc i\kern-.025em b}\kern-.08em
    T\kern-.1667em\lower.7ex\hbox{E}\kern-.125emX}}
\definecolor{abstractbg}{rgb}{1,1,1}

\begin{document}
\title{Concept Drift Challenge in Multimedia Anomaly Detection: A Case Study with Facial Datasets}



\author{Pratibha Kumari, Priyankar Choudhary, Pradeep K. Atrey, and Mukesh Saini 
%
}
\IEEEtitleabstractindextext{%
\fcolorbox{abstractbg}{abstractbg}{%
 \begin{minipage}{\textwidth}%

\begin{abstract}
Anomaly detection in multimedia datasets is a widely studied area. Yet, the concept drift challenge in data has been ignored or poorly handled by the majority of the anomaly detection frameworks. The state-of-the-art approaches assume that the data distribution at training and deployment time will be the same. However, due to various real-life environmental factors, the data may encounter drift in its distribution or can drift from one class to another in the late future. Thus, a one-time trained model might not perform adequately. In this paper, we systematically investigate the effect of concept drift on various detection models and propose a modified Adaptive Gaussian Mixture Model (AGMM) based framework for anomaly detection in multimedia data. In contrast to the baseline AGMM, the proposed extension of AGMM remembers the past for a longer period in order to handle the drift better. Extensive experimental analysis shows that the proposed model better handles the drift in data as compared with the baseline AGMM. Further, to facilitate research and comparison with the proposed framework, we contribute three multimedia datasets constituting faces as samples. The face samples of individuals correspond to the age difference of more than ten years to incorporate a longer temporal context.
\end{abstract}
\end{minipage}
}}
\maketitle
\section{Introduction}\label{sec:intro}
Anomaly detection in multimedia (audio, video, or audio-visual) data has been an active research field for decades due to complex real-world challenges such as unavailability of anomaly samples, high intra-class variance, the vague boundary between anomaly and normal classes, context-dependency, concept drift, etc.~\cite{pang2021deep}. The majority of the challenges are well explored by state-of-the-art deep learning-based unsupervised anomaly detection frameworks. However, concept drift challenge has been mainly ignored while developing the anomaly detection frameworks ~\cite{kumari2020multivariate}. Concept drift refers to the change in the data distribution over time in dynamically changing and non-stationary environments~\cite{widmer1996learning}. A real-life surveillance scene is prone to encounter concept drift which is also reflected in associated audio, video, or any other sensory data. For example, the background sound in an office changes over the day, which reflects a slow drift in the normal data; however, if a background sound observed at night occurs during the day, then it will be an anomaly to the day operator. This implies that the class of a sample may change according to the temporal context information. Another example of drift is as follows: if playing music is prohibited during lunch hours in the office, it is an anomaly; however, if this event occurs frequently, then it might be considered a normal event thereafter. Thus there can be a change in the class of a sample due to various factors.

The majority of recent deep learning-based multimedia anomaly detection frameworks perform competitively on available public datasets which do not contain samples with drifts. However, in the presence of drift which is prevalent in streaming multimedia data, a one-time trained model might not perform well and might need retraining. Alternatively, a lifelong learning paradigm, i.e., adaptive learning, is preferred in various domains of monitoring, such as smart-home~\cite{rashidi2009keeping}, production industry~\cite{pechenizkiy2010online}, telecommunication~\cite{hilas2009designing}, etc.

In the case of streaming multimedia data, AGMM has been majorly used to facilitate adaptive learning for event detection~\cite{cristani2004line,cristani2007audio} and anomaly detection~\cite{kumari2020multivariate,kumari2021situational} tasks. However, AGMM put a constraint on the maximum number of allowed Gaussians (modes) in the mixture. This can cause frequent deletion of recent past in a highly dynamic environment as well as deletion of longer past in a less dynamic environment.

In this paper, we propose a modified AGMM-based anomaly detection framework. Contrary to the baseline AGMM, the proposed extension, namely Unconstrained AGMM (UAGMM), does not set any limit on the maximum possible number of modes in the mixture. Thus, no mode is permanently deleted from the memory. Further, an adaptive learning approach involving AGMM is prone to generate multiple modes for the same unique sample during initial encounters with new samples in the dynamic environment~\cite{moncrieff2007online}. To reduce such modes and hence false classification, a merging facility that merges closer modes is also added in UAGMM.

There is a lack of suitable multimedia datasets for the evaluation of adaptive anomaly detection frameworks. A dataset to be used for evaluation should have the following properties:

(i) Samples should be collected in a temporal sequence.

(ii) Anomalies are contextual. An event or object may be an anomaly in some context, whereas it can be normal in another context. Therefore, the recording device should surveil only a fixed place, activity, scene, or object.

(iii) The dataset should span enough duration to capture the drift in data.

Keeping in mind all the above necessary conditions, we collect three large datasets for evaluation. We opt for collecting face samples of young individuals as they are prone to get the drift in their facial structure when observed for longer years. Though we evaluate our work with face images, it can be used for any multimedia data-based surveillance application.

Concept drift challenges of anomaly detection targeted in this paper is discussed below in detail with examples. 
 
\begin{figure}
\centering
\includegraphics[width=1\linewidth]{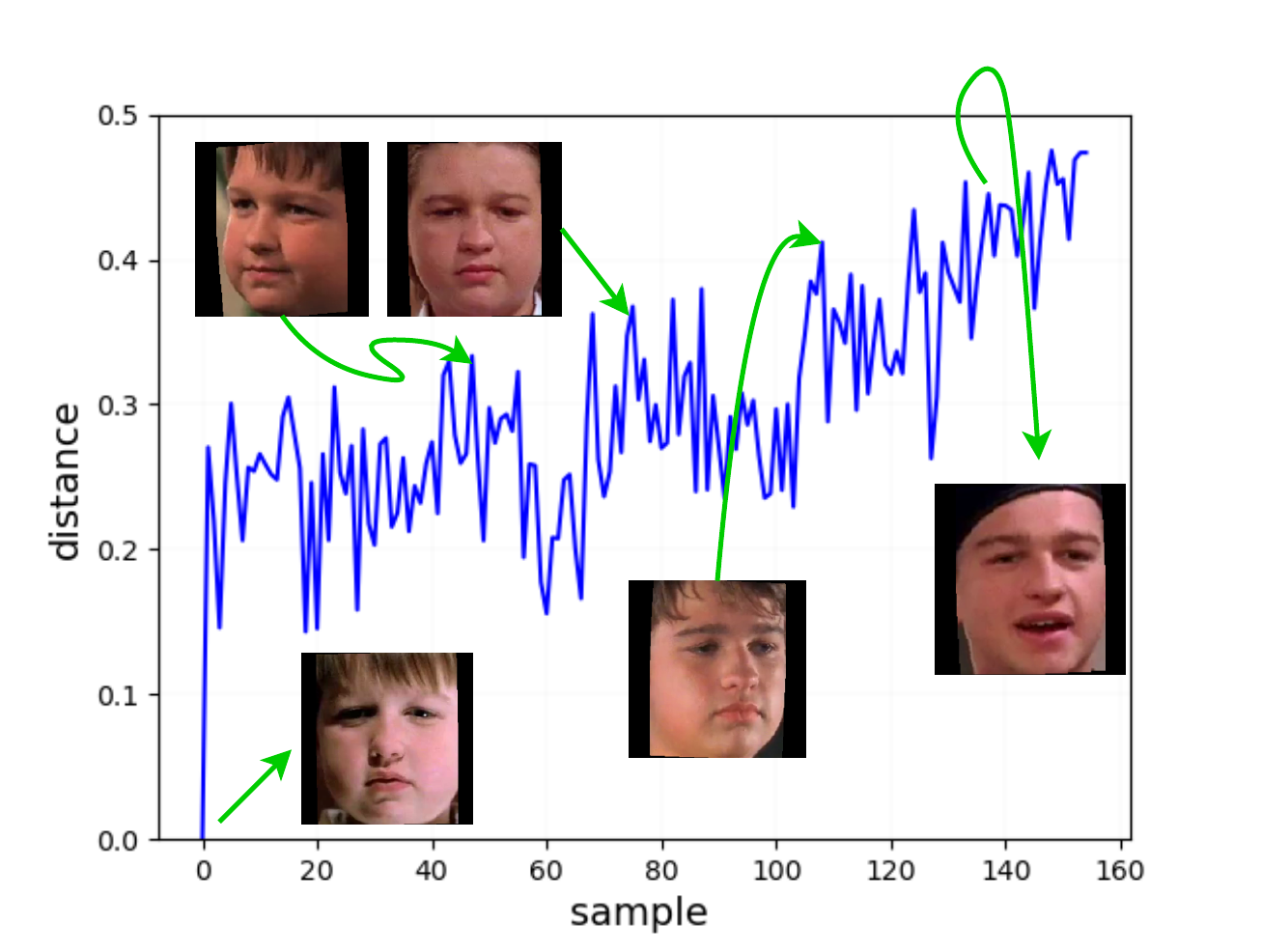}
\caption{\textbf{Drift in the normal data}: figure shows distance of a reference face image with subsequent face images of `Jake'. }
\label{fig:jakeUTubeNotMix}
\end{figure}
\subsection{Concept drift Challenge}
Concept drift in data can lead to the following two drifts: (i) intra-class drift and (ii) inter-class drift. To elaborate, we take a use case of face-based authentication in a smart home where all the appliances work with face recognition and verification. Another application can be face-based authentication in a smart office where the entry gates and appliances work with face recognition and verification. However, the above challenges hold true to all real-life use cases of anomaly detection.
\\\textbf{Intra-class drift}: Intra-class drift corresponds to slow drift within the class, mainly the normal class. In most real-life cases, the data distribution for the normal class is not constant; it changes over time. 
For example, the average yearly expenses of an individual, the population level in a city, the visual appearance of a city from the top, etc., might slowly change over time. The smooth change does not correspond to the presence of anomalies. They are the slow drift in the normal data distribution itself.

For the smart home use case, let's assume that only one person, namely `Jake Harper' (a character from `Two and a Half Men' TV-series), age 8 years, lives in the house. All the smart appliances in the house are able to verify him. But as he grows older, his facial structure changes. Then it might be the case that smart machines will no longer be authenticated with his face. We show some representative frames of `Jake Harper' over the years via Figure~\ref{fig:jakeUTubeNotMix}. In the figure, the x-axis represents face samples taken over time. We compute the cosine distance from a deep embedding of $1^{st}$ sample to the deep embedding of all the subsequent samples and plot. For the $1^{st}$ sample, the distance will be $0$ as it is compared against the same image. We can observe that it shows a trend of increase in distance over time. The state-of-the-art deep face verification models such as `DeepFace'~\cite{serengil2021lightface} keep a static threshold for face verification. With the presented use case, the verification will fail as the model is not adaptive. Additionally, most of the state-of-the-art approaches for anomaly detection are evaluated on short-term data and hence ignore the presence of slow drift in data. To cope with such drift, the model should follow an adaptive learning mechanism.
\\\textbf{Inter-class drift}:
Inter-class drift refers to drift across classes. In a real-world dynamic environment, the frequency of occurrence of a sample may change over time which leads to change in class from abnormal to normal and vice-versa. For example, a birthday celebration in a lab is rare and hence an anomaly. However, if it starts occurring often, then it will no longer be regarded as an anomaly. The change from abnormal class to normal class is regarded as new normality.

For the smart home use case, if some guests arrive and try to use the smart appliances, they will not be verified as their face is encountered less often and hence are regarded as anomalies.

Further, if a guest decides to stay in the house for a longer duration, then he/she should no longer be treated as a guest after a few days. 
It means his face must be regarded as a new normal by the smart appliances after seeing enough occurrences. Thus, we see a drift from abnormal to normal class for a distinct sample. Traditionally, state-of-the-art deep learning-based approaches target to model only the normal class; hence they can miss this drift. For learning this drift, the model should also consider modeling the anomaly class so that any drift in them can also be tackled. Further, if someday `Jake Harper' moves out of the house permanently or reduces his frequency of visiting the home, it becomes doubtful whether he belongs to that home or not. This can cause a shift from normal to abnormal for `Jake Harper' and turn his appearance into an anomaly.
\subsection{Contributions}
We brief our contributions as follows: 
\begin{itemize}

    \item We contribute three facial image-based datasets\footnote{Datasets will be released in the final version of the manuscript} for the evaluation of adaptive anomaly detection frameworks. Face samples of individuals with an age difference of more than ten years so as to incorporate drift are considered.

    \item The proposed extension of baseline AGMM, namely UAGMM, outperforms the previous works and needs less prior knowledge. Further, we give a comparison of various adaptive learning and static learning framework for anomaly detection on the collected datasets.
    
\end{itemize}
Rest of the paper is structured as follows. We do a literature survey in Section~\ref{sec:literature}. Details of dataset collection are provided in Section~\ref{sec:dataset}. We present the proposed approach in Section~\ref{sec:proposed}. Experimental results \& analysis are presented in Section~\ref{sec:exp}. We write the conclusion in Section~\ref{sec:conc}.
\begin{table*}[!htbp]
\centering
\caption{Comparing the state-of-the-art works for anomaly detection in multimedia data with the proposed work}
 \label{tab:comp_literature}
\begin{tabular}{|c|c|c|c|c|c|c|c|c|}
\hline
 Work& \begin{tabular}[c]{@{}c@{}}Research\\ application \end{tabular}  &  \begin{tabular}[c]{@{}c@{}}Learning\\ \& model \\type \end{tabular} &
 \begin{tabular}[c]{@{}c@{}} Dynamic \\ anomaly\\types? \end{tabular}  &\begin{tabular}[c]{@{}c@{}}Concept\\ drift \end{tabular}  &\begin{tabular}[c]{@{}c@{}}Temporal\\ sequence\\matter? \end{tabular} &\begin{tabular}[c]{@{}c@{}}dynamic \\memory \\limit, if\\adaptive? \end{tabular}  &
 \begin{tabular}[c]{@{}c@{}}Non-monotonically \\increasing \\memory, if\\ adaptive? \end{tabular} &\begin{tabular}[c]{@{}c@{}}Model is\\ updated \\after each\\ sample, if\\ adaptive? \end{tabular} 
 \\\hline

 Foggia et al.~\cite{foggia2015audio}& \begin{tabular}[c]{@{}c@{}}road \\surveillance\end{tabular} &  \begin{tabular}[c]{@{}c@{}}static \& \\supervised\end{tabular} &no&none&no&--&--&--\\\hline
 
 \begin{tabular}[c]{@{}c@{}}Foggia et al.~\cite{foggia2015reliable},
\\ Conte et al.~\cite{conte2012ensemble},\\
 Carletti et al.~\cite{carletti2013audio}\end{tabular}&\begin{tabular}[c]{@{}c@{}}scene \\monitoring\end{tabular} & 
 \begin{tabular}[c]{@{}c@{}}static \& \\supervised\end{tabular} &no&none&no&--&--&--\\\hline
 
 Rushe et al.~\cite{rushe2019anomaly}&\begin{tabular}[c]{@{}c@{}}scene \\monitoring\end{tabular} & \begin{tabular}[c]{@{}c@{}} static \& semi-\\supervised\end{tabular}&no&none&no&--&--&--\\\hline
 
 \begin{tabular}[c]{@{}c@{}}Kawaguchi et al.\\~\cite{kawaguchi2021description} DCASE2021\end{tabular}&\begin{tabular}[c]{@{}c@{}}machine \\monitoring\end{tabular} &\begin{tabular}[c]{@{}c@{}} static \& supervised,\\static \& semi-supervised,\\static \& unsupervised\end{tabular}  &no&intra-class&no&--&--&--\\\hline

 \begin{tabular}[c]{@{}c@{}}
 Wang et al.~\cite{wang2018detecting},\\
Hu et al.~\cite{hu2019efficient}\\
 Wang et al.~\cite{wang2020robust},\\
 Ye et al.~\cite{ye2019anopcn},\\
 Park et al.~\cite{park2022fastano}\end{tabular}&\begin{tabular}[c]{@{}c@{}}scene \\monitoring\end{tabular} 
 & \begin{tabular}[c]{@{}c@{}}static \& \\unsupervised\end{tabular} &yes&none&no&--&--&--\\\hline

 \begin{tabular}[c]{@{}c@{}}Kumari and\\ Saini~\cite{kumari2020multivariate}\end{tabular}&\begin{tabular}[c]{@{}c@{}}scene \\monitoring \end{tabular} &
 \begin{tabular}[c]{@{}c@{}}adaptive \& \\unsupervised\end{tabular} 
&yes&\begin{tabular}[c]{@{}c@{}}intra-class,\\inter-class \end{tabular}  &yes&no&yes&yes\\\hline

\begin{tabular}[c]{@{}c@{}}
Id et al.~\cite{id2020concept},
\\Id et al.~\cite{id2022concept}\end{tabular}
&\begin{tabular}[c]{@{}c@{}}scene \\classification\end{tabular} & \begin{tabular}[c]{@{}c@{}}adaptive \& \\unsupervised\end{tabular} &yes&intra-class&yes&yes&no&no\\\hline

Proposed&\begin{tabular}[c]{@{}c@{}}scene\\ monitoring \end{tabular} & \begin{tabular}[c]{@{}c@{}}adaptive \& \\unsupervised\end{tabular} &yes&\begin{tabular}[c]{@{}c@{}}intra-class,\\inter-class \end{tabular}  &yes&yes&yes&yes\\\hline

\end{tabular}
\end{table*}
\section{Literature}\label{sec:literature}
Anomaly detection in multimedia data has been extensively studied in past decades. While early approaches focused on hand-crafted feature-based statistical or machine learning models, state-of-the-art approaches are dominated by deep learning models~\cite{wang2018detecting,lai2020video}. Further, a shift from supervised to semi-supervised or unsupervised learning-based approaches has been witnessed in this area.

In the case of audio-based anomaly detection, research has mainly focused on detecting application-specific anomaly types~\cite{crocco2016audio}. 
For example, car crashes, accidents, and tire-skidding are considered for road surveillance~\cite{foggia2015audio}, glass break, gunshot, baby cry, and shouting are considered for scene monitoring~\cite{foggia2015reliable,conte2012ensemble,carletti2013audio,rushe2019anomaly}, etc. Popular approaches for the above applications include bag-of-words with multiple SVM-based supervised approaches~\cite{carletti2013audio,foggia2015audio}, convolutional autoencoder-based semi-supervised approach~\cite{rushe2019anomaly}, etc.
Another popular application is monitoring industrial machines by observing the sound produced by them.
Abnormality in a machine's normal sound may indicate potential damage to the machine. DCASE challenge is the widely accepted venue for machine monitoring datasets. Domain shift has been introduced in the latest challenges there~\cite{kawaguchi2021description}. Implying that the recording environment of sound samples is changed from train to test as a condition of weather and machines are different at these times. The temporal order of events has no importance in these datasets, whereas the intra-class and inter-class drifts considered in this paper depend upon the temporal sequence of occurrences of activities/ events. Consequently, the drift being tackled in the proposed work can not be covered in those datasets. Hence, there has not been any development in adaptive anomaly detection approaches for streaming audio data.

Video anomaly detection approaches have been largely developed for indoor and outdoor scene monitoring. In contrast to audio-based anomaly detection, generally, heterogeneous anomaly types are considered in video anomaly detection. However, available video anomaly datasets have a stationary data distribution.

State-of-the-art unsupervised video anomaly detection techniques are predominantly based on applying reconstruction-based approaches~\cite{wang2018detecting,hu2019efficient} or prediction-based approaches~\cite{wang2020robust,ye2019anopcn,park2022fastano}. 
The reconstruction-based approaches aim to reconstruct the present input frame and assign an anomaly score based on the reconstruction error. Whereas prediction-based approaches aim at correctly predicting the immediate future frame, and the prediction error is used to assign an anomaly score. Generally, these models are trained once using samples from normal class and deployed as a static model. Since the data used to train these models is too short and discontinuous, hence they do not contain samples with concept drift. These models always classify a sample as abnormal during the test phase in case it was classified as abnormal during the training phase, regardless of how frequently it occurs in the distant future~\cite{kumari2021situational}. While these models perform competitively on the public datasets, their performance on a streaming dataset will be inadequate.

\begin{table*}[!h]
\centering
\caption{Datasets comparison}
 \label{tab:comp_data}
\begin{tabular}{|c|c|c|c|c|c|}
\hline
 Dataset Reference&Modality&\begin{tabular}[c]{@{}c@{}}Single\\ scene?\end{tabular} &\begin{tabular}[c]{@{}c@{}}Streaming\\ data?\end{tabular} & Time-span&\begin{tabular}[c]{@{}c@{}}Concept drift\\ considered in data \end{tabular}  
 \\\hline

\begin{tabular}[c]{@{}c@{}} Human-human \\interaction~\cite{lefter2014audio}\end{tabular} &audio-visual&yes&no &$\approx$33 minutes&none  \\
VSD~\cite{demarty2015vsd} &audio-visual&no&no &$\approx$ 29 hours   &none \\
EMOLY~\cite{fayet2018emo} &audio-visual&yes&no &-- &none \\
XD-Violence~\cite{wu2020not} &audio-visual&no&no &217 hours&none \\

DCASE2017~\cite{DCASE2017}&audio&no&no &39 minutes   &none  \\
DCASE2021~\cite{kawaguchi2021description} &audio&no&no & $\approx$82 hours  &intra-class \\

UCSD Ped 1~\cite{mahadevan2010anomaly} &visual&yes&no &$\approx$ 5-7 minutes  &none \\
UCSD Ped 2~\cite{mahadevan2010anomaly} &visual&yes&no &$\approx$ 2-3 minutes  &none  \\

UMN~\cite{xyz} &visual&no&no &$\approx$5 minutes  &none  \\
AVENUE~\cite{lu2013abnormal}  &visual&no&no &$\approx$20 minutes &none  \\
ShanghaiTech~\cite{luo2017revisit}  &visual&no&no &$\approx$3-3.5 hours  &none  \\
Street Scene~\cite{ramachandra2020street}  &visual&yes&no &$\approx$1-2 year  &none \\
UCF-Crime~\cite{sultani2018real} &visual&no&no &$\approx$ 128 hours &none \\\hline

Subway Entrance~\cite{adam2008robust} &visual&yes&yes &$\approx$ 96 minutes  &none  \\
Subway Exit~\cite{adam2008robust} &visual&yes&yes &$\approx$ 43 minutes&none  \\
ADOC~\cite{pranav2020day}  &visual&yes&yes &1 day &none\\\hline

Proposed: JakeFDB&visual&yes&yes &$\approx$11 years &\begin{tabular}[c]{@{}c@{}}intra-class, inter-class \end{tabular}  \\
Proposed: DebbieFDB&visual&yes&yes &$\approx$10 years &\begin{tabular}[c]{@{}c@{}}intra-class, inter-class \end{tabular}  \\
Proposed: HarryFDB&visual&yes&yes &$\approx$11 years &\begin{tabular}[c]{@{}c@{}}intra-class, inter-class \end{tabular}  \\\hline

\end{tabular}
\end{table*}

Id et al.~\cite{id2020concept,id2022concept} employed GMM based adaptive framework, namely Combine-merge Gaussian Mixture Model (CMGMM), for acoustic scene classification. The approach facilitates combining and merging the existing mixture with a new mixture, computed from drifted data to handle concept drift. Initially, GMMs with different K (maximum number of allowed Gaussians in the GMM) are trained using the expectation–maximization (EM) algorithm, and the best GMM is selected using the Bayesian information criterion (BIC). They used kernel density estimation to detect drift in data windows. If drift is found in some data windows, a new mixture is trained using EM and merged with the older mixture. Contrary to AGMM-based approaches where the mixture is updated after each sample, in CMGMM, the update is performed after a window of data. Thus finding a suitable window size for a surveillance environment becomes a bottleneck in this approach. 
Further, the number of Gaussians in the merged mixture has to be at least equal to those in the older mixture. Therefore with this strategy, the number of modes in the mixture monotonically increases over time. Further, the approach is computationally costly to be used for streaming data~\cite{id2022concept}.  
Kumari and Saini~\cite{kumari2020multivariate} used AGMM based adaptive framework for anomaly detection in an untrimmed video dataset. Although the model adapts to the changing data distribution, it frequently forgets the distant past due to the frequent replacement of weak Gaussians in the mixture. 
In contrast, we present drift-based challenges in anomaly detection, its analysis, and a baseline approach in a more comprehensive manner.

A comparison of the existing works on anomaly detection with the proposed work is presented in Table~\ref{tab:comp_literature}.
The comparison is provided from the following aspects: (1) what type of learning is employed, i.e., static or adaptive, and whether the model is supervised, semi-supervised, or unsupervised, (2) whether dynamic types of anomalies or a few fixed classes of anomaly are only considered, (3) whether concept drift in data is considered or not; if considered then what are the types of concept drifts (4) does the data need to follow a temporal sequence, i.e., it should be streaming or not, (5) if the model follows adaptive learning, then whether the upper limit on memory is dynamic, (6) if the model follows adaptive learning, then the model’s memory dynamically adjust itself or monotonically increases, and (7) if the model follows adaptive learning, then the model's parameters are updated after each sample or a window of samples. It is evident from the table that the proposed work is novel in many aspects. There have been very few attempts toward anomaly detection in streaming multimedia data. We are the first to tackle intra-class and inter-class drifts with no limitation on the model's memory. Further, while ensuring dynamic memory, the model does not cause a monotonic increment in memory over time. Contrary to the approach by Id et al.~\cite{id2020concept,id2022concept}, the changes are reflected after each sample which removes the need to find a suitable window length.
\begin{figure*}[!htbp]
\centering
\includegraphics[scale=0.67]{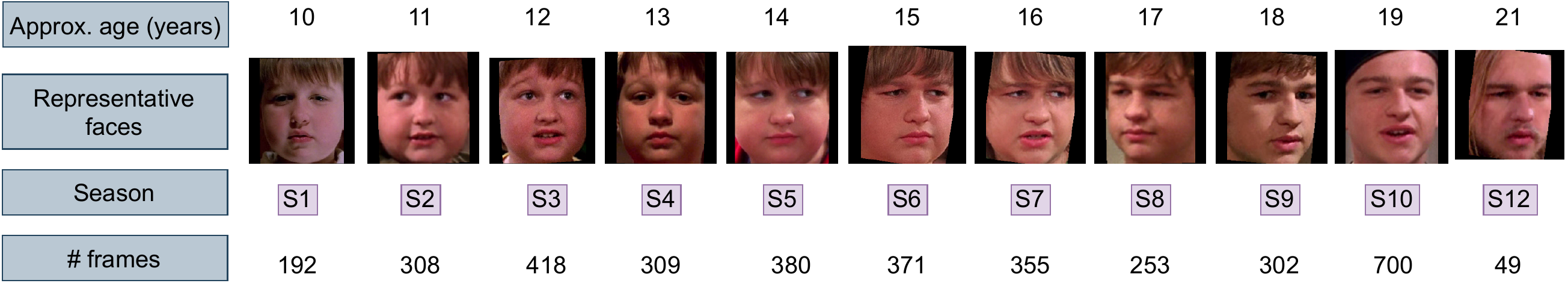}
\caption{Representative face samples of `Jake Harper' from different seasons in JakeFDB dataset}
\label{fig:jakeOverSeason}
\end{figure*}
\begin{figure*}[!htbp]
\centering
\includegraphics[scale=0.67]{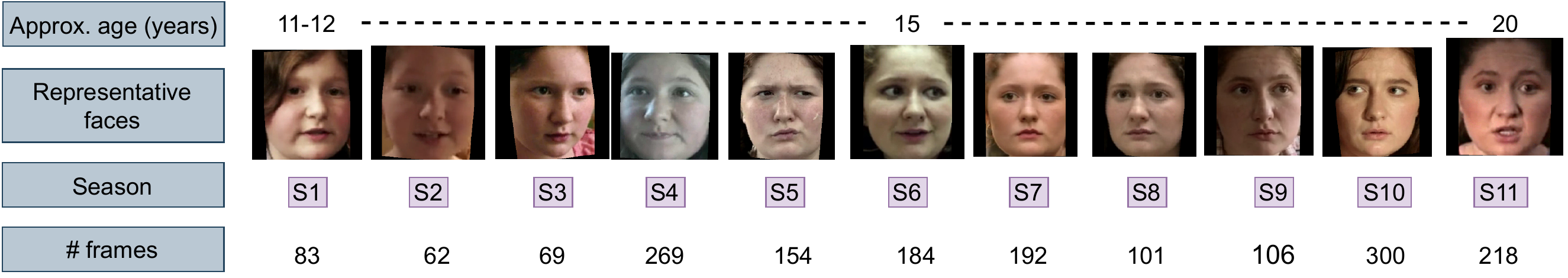}
\caption{Representative face samples of `Debbie Gallagher' from different seasons in DebbieFDB dataset}
\label{fig:debyOverSeason}
\end{figure*}
\section{Dataset}\label{sec:dataset}
A multimedia dataset that is created by surveilling a single scene or object or situation in a continuous manner for a long duration enough to capture drift can be used for the evaluation of adaptive anomaly detection frameworks. Most of the existing multimedia datasets do not fulfill all of the above criteria. Therefore, we collect three face image-based datasets for the evaluation of adaptive anomaly detection frameworks. We list a comparison of popular multimedia datasets for anomaly detection with the proposed datasets in Table~\ref{tab:comp_data}. It can be seen that most of the datasets are recorded in non-streaming and multi-scene setups. Few datasets, viz., Subway Entrance, Subway Exit, and ADOC are streaming and single-scene, but they are not initially developed for anomaly detection under concept drift situations, and hence concept drift is not focused in these datasets. Moreover, they are of a very short duration compared to the proposed datasets.

For dataset collection, we consider human faces as samples. However, collecting a dataset that comprises of faces of an individual over a large range of years is challenging as it requires a longer waiting time to get sufficient samples. Alternatively, we can look into stored CCTV footage to get samples for a specific person. But, generally, the surveillance footage is not stored for years due to space constraints and hence is deleted periodically. 
On the other hand, some TV-series or movies have featured a few characters over a longer range of their age. We found that collecting samples using such TV shows will be suitable for our purpose. Moreover, the drift in one's face is prevalent more during the transition from childhood to adulthood age. Therefore, we pick a few TV-series and movie-series where a child actor is cast in his/her teenage and featured till his/ her adult age. Three datasets are collected for model evaluation purposes. We opted for two TV-series, viz., `Two and a Half Men', `Shameless' and one movie-series, viz., `Harry Potter', to create datasets.

From the `Two and a Half Men' TV-series, we considered a male character, namely `Jake Harper' to collect the first dataset, namely JakeFDB. This TV-series has a total of 12 seasons in which `Jake Harper' has appeared in all seasons except season-11. Also, his appearance in season-12 was as a guest and hence, very few occurrences as compared with other seasons. The choice of this TV-series is such that it covers a visible change in the facial structure of `Jake Harper'. The change in `Jake Harper' over the years can also be observed in Figure~\ref{fig:jakeOverSeason}. We show representative face images from each season along with his approximate age during a particular season. He was 10 years old at his first appearance and 21 years old at his last appearance in the show. Our dataset is made by scrapping the faces of `Jake Harper' from the top 5 episodes from each season (except for the season-11) of `Two and a Half Men'. The season-wise number of samples of `Jake Harper’ in the proposed dataset JakeFDB is also listed in Figure~\ref{fig:jakeOverSeason}.

The second dataset, DebbieFDB, is collected from the ‘Shameless’ TV-series. 
A female character, namely Debbie Gallagher’ from this TV-series, is considered for the creation of the DebbieFDB dataset. The ‘Shameless’ TV-series has a total of 11 seasons in which ‘Debbie Gallagher’ appeared in all seasons with varying frequency. Her appearance in the first season was very less; however, it increased as the season advanced. ‘Debbie Gallagher’ was 11-12 years old at her first appearance and 20 years old at her last appearance at the show. The changes in her face over the seasons can be seen in Figure~\ref{fig:debyOverSeason}. Her age during each season is not made public; however, her age in a few seasons is available, viz., S1, S6, and S11. The top 5-7 episodes from each season are considered to create this dataset. Figure~\ref{fig:debyOverSeason} also shows the season-wise number of samples of ‘Debbie Gallagher’ in the DebbieFDB dataset.

\begin{figure*}[!htbp]
\centering
\includegraphics[scale=0.67]{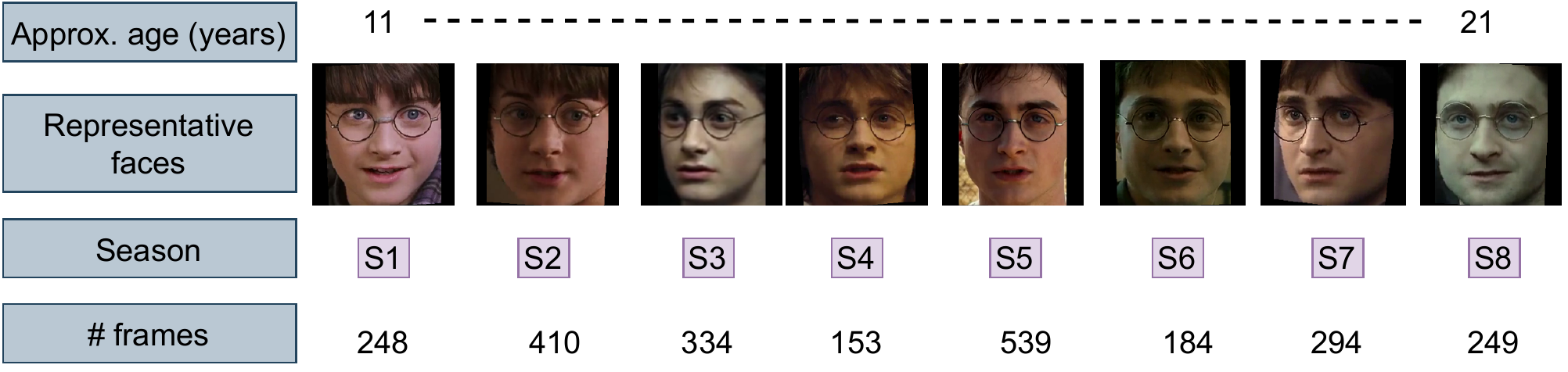}
\caption{Representative face samples of `Harry Potter' from different seasons in HarryFDB dataset}
\label{fig:harryOverSeason}
\end{figure*}

The third dataset, namely HarryFDB, is collected using all the parts of the `Harry Potter' movie-series. For naming convenience, we call the 1st movie of the `Harry Potter' series as season-1 and so on. For HarryFDB, we consider a male child character named `Harry Potter' from these movies. This movie-series has 8 parts in which `Harry Potter' appeared in all parts as the lead character. Thus, we get a similar number of samples from all the parts. We show the representative face samples for `Harry Potter' over the seasons via Figure~\ref{fig:harryOverSeason}. He was 12 years old at his first appearance and 22 at his last appearance in the movie-series. Each movie part is shot at a different interval of his age. However, they are in temporal sequence order and hence suitable for analysis.

\begin{table*}[!htbp]
\centering
\caption{Details of video clips used for creation of proposed datasets }
 \label{tab:FDB_Description}
\begin{tabular}{|c|c|c|c|}
\hline
 Dataset$\longrightarrow$&JakeFDB&DebbieFDB& HarryFDB \\
 \hline
Name of the TV-series/ movie-series &Tow and Half Men&Shameless&Harry Potter\\\hline 
Total \# seasons&11&11&8\\\hline
\# of episodes taken from each season& 5& 5-7&whole\\\hline
Duration for each season (minutes)& 5*24& (5-7)*(44-60)&(130-161)\\\hline
FPS& 1& 1&1\\\hline
Resolution& 1920*1080& 1920*1200&$\approx$1280*528\\\hline
\end{tabular}
\end{table*}
\begin{figure}[!htbp]
\begin{minipage}[b]{.19\linewidth}
  \centering
 \includegraphics[width = 1\linewidth]{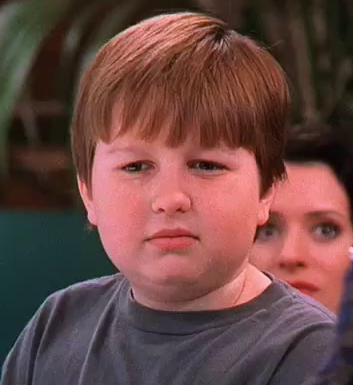}
\end{minipage}
\hfill
\begin{minipage}[b]{.19\linewidth}
  \centering
 \includegraphics[width = 1\linewidth]{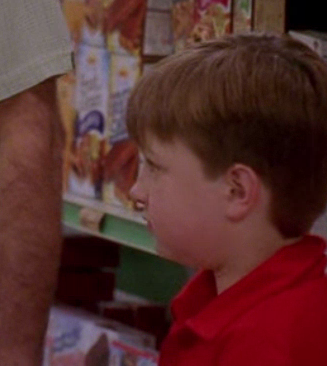}
\end{minipage}
\hfill
\begin{minipage}[b]{.19\linewidth}
  \centering
 \includegraphics[width = 1\linewidth]{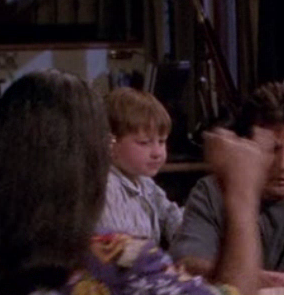}
\end{minipage}
\hfill
\begin{minipage}[b]{.19\linewidth}
  \centering
 \includegraphics[width = 1\linewidth]{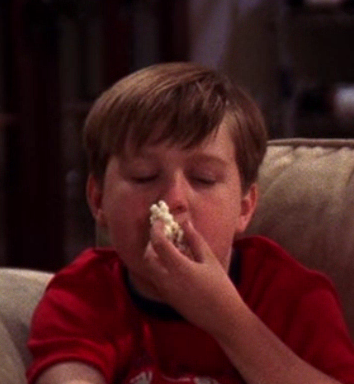}
\end{minipage}
\hfill
\begin{minipage}[b]{.19\linewidth}
  \centering
 \includegraphics[width = 1\linewidth]{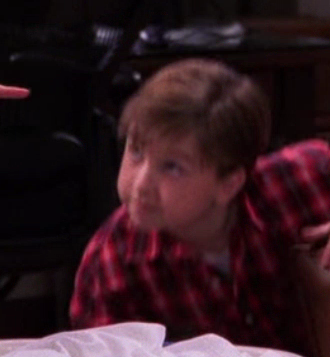}
\end{minipage}
\begin{minipage}[b]{.19\linewidth}
  \centering
 \includegraphics[width = 1\linewidth]{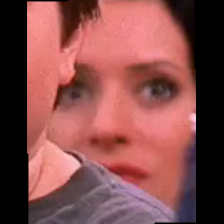}
 \centerline{other face}
\end{minipage}
\hfill
\begin{minipage}[b]{.19\linewidth}
  \centering
 \includegraphics[width = 1\linewidth]{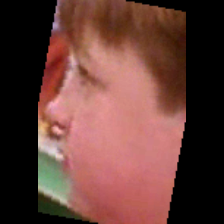}
 \centerline{side face}
\end{minipage}
\hfill
\begin{minipage}[b]{.19\linewidth}
  \centering
 \includegraphics[width = 1\linewidth]{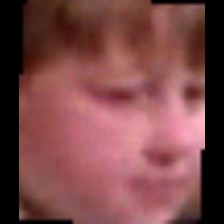}
 \centerline{far distance}
\end{minipage}
\hfill
\begin{minipage}[b]{.19\linewidth}
  \centering
 \includegraphics[width = 1\linewidth]{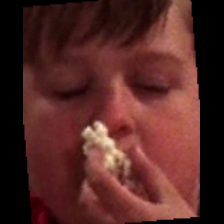}
 \centerline{occlusion}
\end{minipage}
\hfill
\begin{minipage}[b]{.19\linewidth}
  \centering
 \includegraphics[width = 1\linewidth]{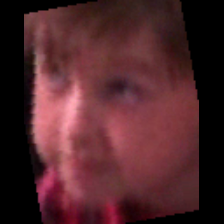}
 \centerline{blur}
\end{minipage}
\caption{Some example of deleted frames from JakeFDB during manual filtering step}
\label{fig:deleted}
\end{figure}

The details of considered TV-series and movie-series to create the three datasets, viz., JakeFDB, DebbieFDB, and HarryFDB are summarized in Table~\ref{tab:FDB_Description}.

The process of creating samples for all datasets are same and majorly automated. For a better understanding, we illustrate the process with the JakeFDB dataset. We list the steps of scrapping samples from a season, e.g., season-1 (S1), as follows:
\begin{itemize}
\item \textbf{Collect all faces from S1}: We take the top 5-7 episodes (each episode is 24 minutes, 1920*1080 resolution, 15 FPS (Frame Per Second) from S1. We extract frames with 1 FPS and detect all the faces from each frame using a state-of-the-art face detector Multi-Task Cascaded Convolutional Neural Networks (MTCNN)~\cite{zhang2016joint}. Then, all these faces are stored with some margin from the top, bottom, left, and right.
Let’s call this database $FDB_{S1}$.

\item \textbf{Filter faces of `Jake Harper' only}: Currently, the database has the faces of all the characters in S1. We need
to take out all the faces of `Jake Harper’. For this, we take 5 faces of `Jake Harper’ as template frames from $FDB_{S1}$. This is performed manually and covers a variety of faces of `Jake Harper’ from S1. We apply face verification using these template faces to all the faces in $FDB_{S1}$. If any of the templates match a face in $FDB_{S1}$, we filter it out. Thus, in the end, we have all
the `Jake Harper’ faces from S1. We say this filtered database as $FDB_{S1}^{'}$.

\item \textbf{Manual filtering}: The filtered database can still have some unwanted faces of ‘Jake Harper’. This includes blurred face, occlusion, side faces, etc. We manually remove such faces from $FDB_{S1}^{'}$. Specifically, we remove all such facial images which have: (a) side view with only one eye visible, (b) face/eyes/mouth occluded by some object, (c) blurred faces, (d) other faces very close, and (e) image taken from far distance. A few examples of such removed images (row-1) with their corresponding output of MTCNN (row-2) are reported in Figure~\ref{fig:deleted}.
\end{itemize}

After following the above steps for each season of the ‘Two and a Half Men’ TV-series separately, we get 192, 308, 418, 309, 380, 371, 355, 253, 302, 700, and 49 samples from S1, S2, S3, S4, S5, S6, S7, S8, S9, S10, and S12, respectively. Thus, the final database JakeFDB has 3637 images, all with ‘Jake Harper’ faces. In the case of DebbieFDB dataset creation from ‘Shameless’ TV-series, we get 83, 62, 69, 269, 154, 184, 192, 101, 106, 300, and 218 samples from S1, S2, S3, S4, S5, S6, S7, S8, S9, S10, and S11, respectively. Thus, we get a total of 1738 samples for DebbieFDB. 
For HarryFDB, instead of a few parts of a season, the entire season of `Harry Potter' movie-series is considered for dataset creation. We get 248, 410, 334, 153, 539, 184, 294, and 249 samples from S1, S2, S3, S4, S5, S6, S7, and S8, respectively. It leads to a total of 2411 samples in HarryFDB. These details are also mentioned in Figure~\ref{fig:jakeOverSeason} to Figure~\ref{fig:harryOverSeason}.
\begin{figure}[!htbp]
\begin{minipage}[b]{.19\linewidth}
  \centering
 \includegraphics[width = 1\linewidth]{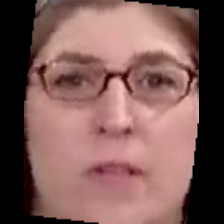}
\end{minipage}
\hfill
\begin{minipage}[b]{.19\linewidth}
  \centering
 \includegraphics[width = 1\linewidth]{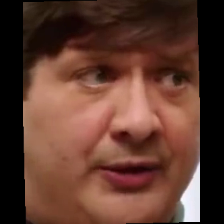}
\end{minipage}
\hfill
\begin{minipage}[b]{.19\linewidth}
  \centering
 \includegraphics[width = 1\linewidth]{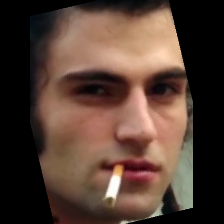}
\end{minipage}
\hfill
\begin{minipage}[b]{.19\linewidth}
  \centering
 \includegraphics[width = 1\linewidth]{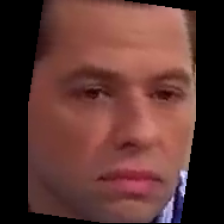}
\end{minipage}
\hfill
\begin{minipage}[b]{.19\linewidth}
  \centering
 \includegraphics[width = 1\linewidth]{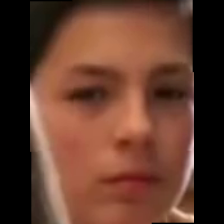}
\end{minipage}
\begin{minipage}[b]{.19\linewidth}
  \centering
 \includegraphics[width = 1\linewidth]{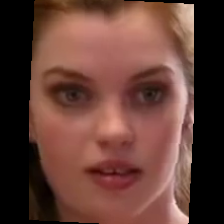}
\end{minipage}
\hfill
\begin{minipage}[b]{.19\linewidth}
  \centering
 \includegraphics[width = 1\linewidth]{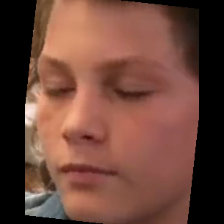}
\end{minipage}
\hfill
\begin{minipage}[b]{.19\linewidth}
  \centering
 \includegraphics[width = 1\linewidth]{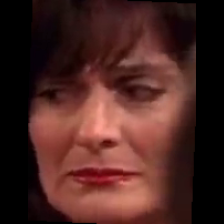}
\end{minipage}
\hfill
\begin{minipage}[b]{.19\linewidth}
  \centering
 \includegraphics[width = 1\linewidth]{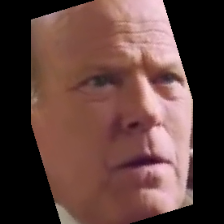}
\end{minipage}
\hfill
\begin{minipage}[b]{.19\linewidth}
  \centering
 \includegraphics[width = 1\linewidth]{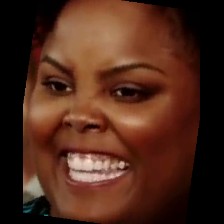}
\end{minipage}
\caption{Faces of other characters taken as anomaly samples in proposed datasets}
\label{fig:sampleAnomaly}
\end{figure}
\begin{figure*}[!htbp]
  \centering
    \includegraphics[scale=0.75]{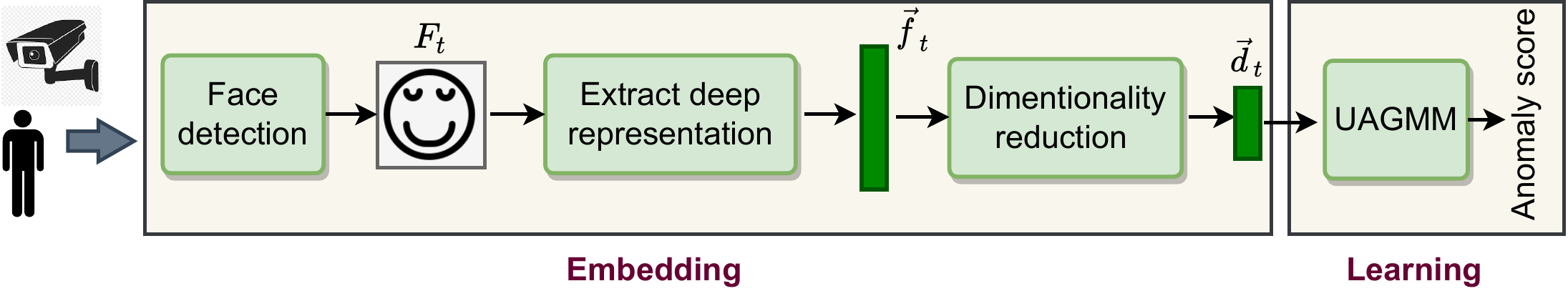}
  \caption{Coarse-level overview of the proposed framework}
  \label{fig:frameworkFace}
\end{figure*}
\\
\textbf{Anomaly Samples in Our Datasets}: 
In the above section, the creation of normal class samples is discussed. Here, details about the abnormal samples in our database are provided. We consider the faces of a few characters from `Two and a Half Men’ and `Young Sheldon’ TV-series as anomaly faces. We take two short clips showing the trailer for these TV-series from YouTube\footnote{\href{https://www.youtube.com/watch?v=TEJiXnMuLcs&ab_channel=WaitaMinute}{Anomaly source link 1}}, 
\footnote{\href{https://www.youtube.com/watch?v=iGKnI3Lc7C4&ab_channel=SeriesTrailerMP}{Anomaly source link 2}}. Similar to normal data, we collect frames at 1 FPS and extract faces from each frame using MTCNN. Further, we manually take out the faces of a few characters other than those considered for normal sample creation as anomaly samples. A total of 10 faces are considered anomaly samples in our datasets. These faces are shown in Figure~\ref{fig:sampleAnomaly}.
\section{Proposed work}\label{sec:proposed}
In the context of multi-face authentication-based smart devices, the task of the embedded model is to assign an anomaly score between 0-1 to each encountered face, where a higher value corresponds to a strong indication of anomaly. 
A coarse-level overview of the proposed framework in the above-mentioned context is shown in Figure~\ref{fig:frameworkFace}. It comprises two modules, viz., `Embedding' and `Learning'. Detection of the face, extracting deep representation of the face, and reducing the feature dimensionality are the steps carried out in the ‘Embedding’ module. Further, UAGMM-based adaptive anomaly modeling is performed in the ‘Learning’ module of the framework. We discuss these modules in detail in the following sub-sections.
\subsection{Face Detection and Representation}\label{sec:represent}
Whenever a person goes in front of the camera in the smart appliance, the face detector detects his/her face. For detecting the faces, MTCNN~\cite{zhang2016joint}, a popular deep learning-based model for face detection, is used. Let us denote the face sample at time $t$ as $F_{t}$. Once the face is detected, its deep representation, $\vec{f}_{t}$, is extracted using VGG-Face~\cite{parkhi2015deep}, a widely used deep learning-based model for extracting facial features.

Generally, descriptors extracted from deep learning frameworks are of very high dimensions. Anomaly detection using proximity-based measures, which is generally used in the GMM or clustering-based approaches, performs poorly with high-dimensional features. This is so because in the higher dimension, both normal as well as abnormal becomes sparse, and thus the notion of proximity fails to retain its meaning~\cite{aggarwal2015outlier}. Therefore, a dimensionality reduction step using PCA (Principal Component Analysis)~\cite{tipping1999probabilistic} is carried out.
Specifically, the extracted high-dimensional descriptor $\vec{f}_{t}$ is mapped to lower-dimensional feature vector $\vec{d}_{t}$ using a PCA model. The PCA model is trained using the training data, and the learned model is used for dimensionality reduction at the test time. 
\subsection{Adaptive Learning}\label{sec:faceModeling}
After extracting the deep representation of the face, it is passed to the UAGMM model for the adaptive modeling and anomaly score computation. The proposed UAGMM model is an extension of AGMM. The baseline AGMM algorithm was originally used in~\cite{stauffer1999adaptive} for foreground-background separation in a continuous video stream. Some of the modes (Gaussian) in the mixture correspond to foregrounds and rest to backgrounds. The decision for foreground-background is taken based on a predefined threshold ($G$). First, the modes are sorted based on the supporting samples (known as mode's weight) and variance. The strength of a mode is measured as a percentage of the total number of samples that have previously belonged to that mode. The modes from the top are picked such that they suffice for background with the threshold $G$. Similar to the adaptive foreground-background separation task, the AGMM has been successfully employed for event detection~\cite{cristani2007audio} and anomaly detection~\cite{kumari2020multivariate} tasks.

For multivariate cases, generally, the independence between features and hence a diagonal covariance matrix is considered for simplicity. Whenever a new sample ($\vec{d}_{t}$) arrives, it is matched to each mode using a matching criterion. Similar to~\cite{kumari2020multivariate}, we use Mahalanobis distance to estimate the association of a sample with a mode. If the Mahalanobis distance with the best match (the mode with which the Mahalanobis distance of $\vec{d}_{t}$ is minimum) is found below a threshold ($\theta_{match}$), then it is considered a hit with that mode, otherwise a miss.  

For each $i^{th}$ mode, a weight ($w_{i,t}$) is maintained, which account for the number of time the mode is hit. In order to reflect the inclusion of the new sample in the mixture, the weight of each mode is updated. At time `t', we update the weights according to the following equation:-
\begin{equation}
    w_{i,t+1}=(1-\alpha)w_{i,t}+ \alpha\left ( M_{i,t} \right )
\label{eq:weight}
\end{equation}  

here $M_{i,t}= 1$ if $i^{th}$ mode is matched otherwise 0, $\alpha$ is the model update rate kept between 0 and 1.

Further, in case of hit, the parameters of the hit mode (let its index be $h$) are updated as follows:
\begin{equation}
    \vec{\mu}_{h,t+1}=(1-\beta_{t})\vec{\mu}_{h,t}+ \beta_{t} \vec{d}_{t}
     \label{eq:mu}
\end{equation}
\begin{equation}
    \vec{\sigma}_{h,t+1}^{2}=(1-\beta_{t}) ( \vec{\sigma}_{h,t}^{2})+
 \beta_{t} (\vec{d_{t}} -\vec{\mu}_{h,t+1})^{T} (\vec{d_{t}}-\vec{\mu}_{h,t+1})
  \label{eq:var}
\end{equation}

here, $\vec{\sigma}_{h,t}^{2}$ and $\vec{\mu}_{h,t}$ represent the variance and the mean vector of $h^{th}$ mode at time $t$; and $\vec{\sigma}_{h,t+1}^{2}$ and $\vec{\mu}_{h,t+1}$ at time $t+1$, respectively; $\beta_{t}$ is the distribution update factor inversely proportional to the distance between $\vec{\mu}_{h,t}$ and $\vec{d}_{t}$. The coefficient $\beta_{t}$ is computed as follows:
\begin{equation}
     \beta_{t}=\left (1- \frac{dist(\vec{\mu}_{h,t},\vec{d}_{t})}{\theta_{match}} \right )  \times   \alpha
     \label{eq:gammaupdate}
\end{equation}
here, $dist(\vec{\mu}_{h,t},\vec{d}_{t})$ is the Mahalanobis distance between $\vec{\mu}_{h,t}$ and $\vec{d}_{t}$.

If we get a miss, i.e., if a match is not found in the mixture, a new mode is created corresponding to the new sample. It is initialized with a small initial weight ($w_{0}$), a high variance ($Z$), and the mean vector as the current sample, i.e., $\vec{d}_{t}$.

Traditionally, when the total number of modes in a mixture reaches a predetermined limit ($K$), and a new mode is required to be created, then the weakest mode is replaced by the new mode. But, in our approach, we do not replace a mode but keep including all possible new modes.

After processing a sample, weights need to be normalized to avoid overflow. We normalize weights such that their sum is 1.

The anomaly score ($\Omega$) of the sample $f_{t}$ is computed as $1-w_{\hat{i},t}$ where $\hat{i}$ is the index of matched mode (in case of hit) or newly created mode (in case of a miss).

In online learning, when the actual data distribution is not known in advance, some modes may be wrongly initialized~\cite{moncrieff2007online}. A mode can be initialized in the form of two distinct modes. Therefore, we introduce an iterative mode merge strategy for combining similar modes that are drifting towards each other.  
We use the Bhattacharyya distance measure~\cite{bhattacharyya1943measure} to compute the overlap between two multivariate Gaussians. If the distance between two modes is less than a predefined threshold value ($\theta_{Bhat}$), then they are merged. We continue merging the two most similar modes until no such pairs exist. The parameter of the output mode ($AB$) upon merging the modes $A$ ($\vec{\mu}_{A}$, $\vec{\sigma}_{A}^{2}$) and $B$ ($\vec{\mu}_{B}$, $\vec{\sigma}_{B}^{2}$) with weights $w_A$ and $w_B$ is computed as follows~\cite{park2018fundamentals}:
\begin{equation}
\vec{\mu}_{AB}=\frac{w_A\vec{\mu}_A+w_B\vec{\mu}_B}{w_A+w_B}
\label{eq:mergeeq1}
\end{equation}
\begin{equation}
\vec{\sigma}_{AB}^{2}=\frac{w^2_A}{w^2_A+w^2_B}\vec{\sigma}_{A}^{2} + \frac{w^2_B}{w^2_A+w^2_B}\vec{\sigma}_{B}^{2}
\label{eq:mergeeq2}
\end{equation}
The weight of the output mode is calculated as the sum of the weights of the modes being merged. The merging process reduces redundant modes that may contribute to an increase in false alarms. Additionally, it decreases memory requirements.
\begin{figure*}[!h]
\begin{minipage}[b]{.32\linewidth}
  \centering
 \includegraphics[width = 1\linewidth]{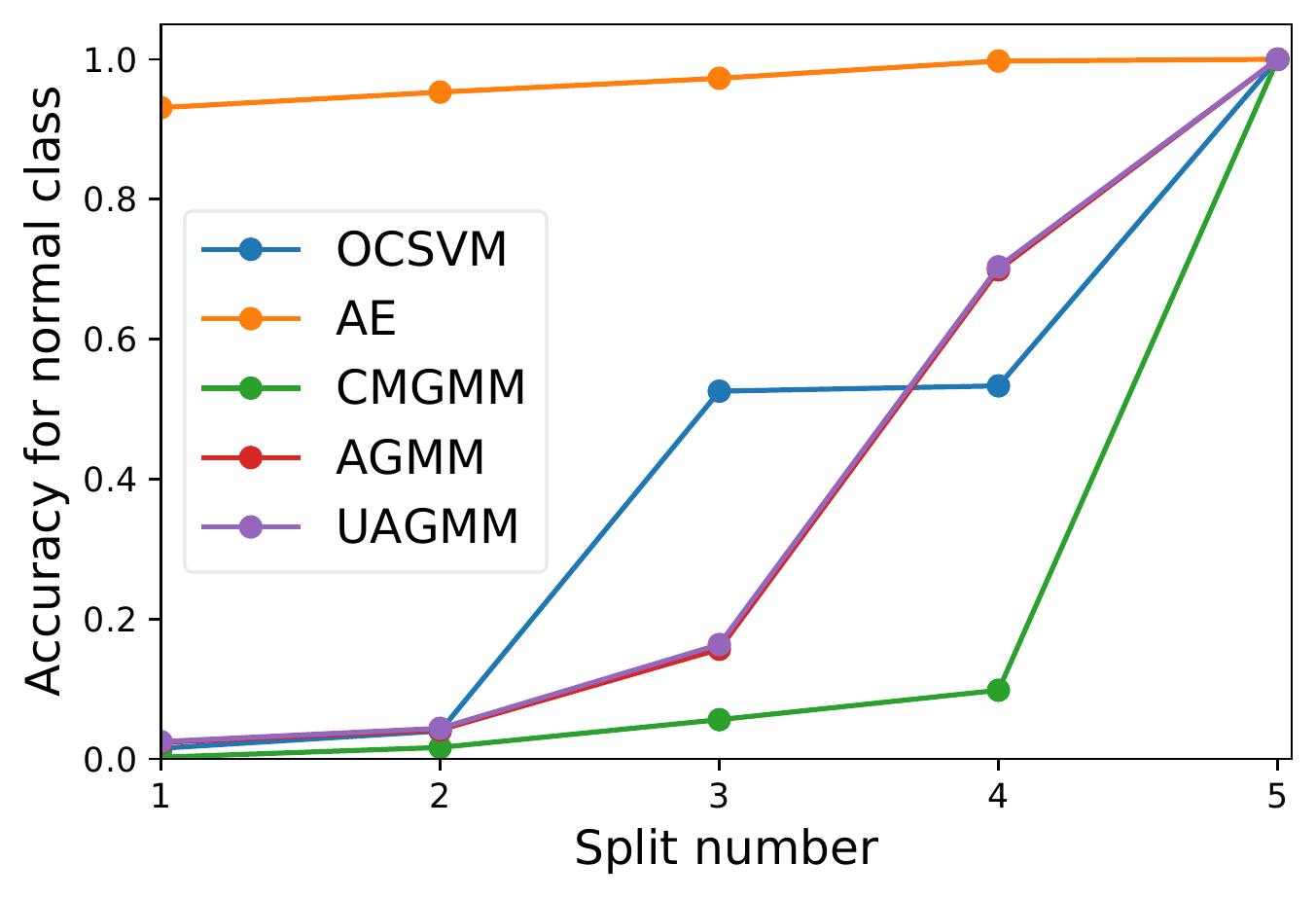}
  \centerline{(a) JakeFDB}\medskip
\end{minipage}
\hfill
\begin{minipage}[b]{.32\linewidth}
  \centering
 \includegraphics[width = 1\linewidth]{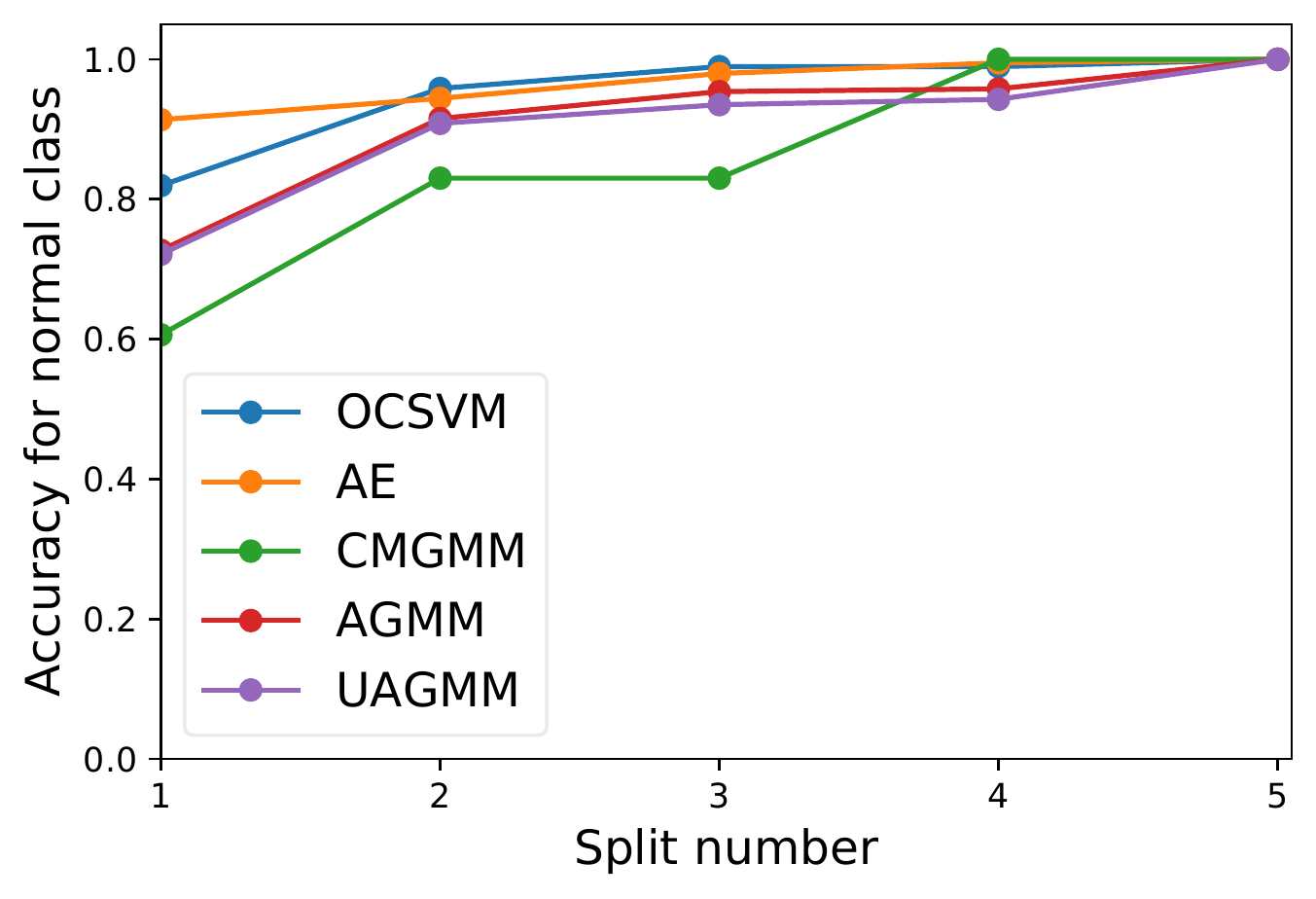}
   \centerline{(b) DebbieFDB}\medskip
\end{minipage}
\hfill
\begin{minipage}[b]{.32\linewidth}
  \centering
 \includegraphics[width = 1\linewidth]{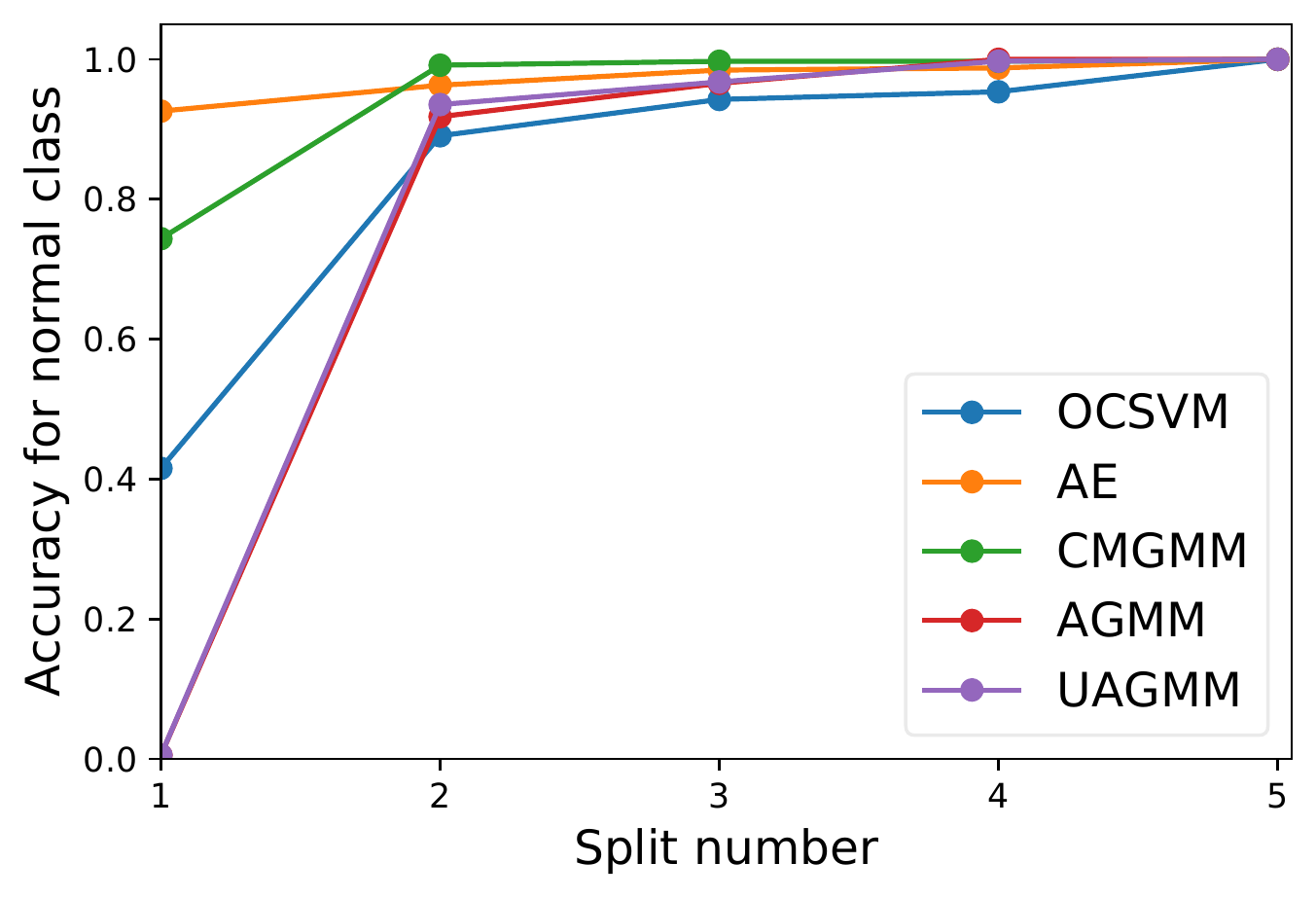}
   \centerline{(b) HarryFDB}\medskip
\end{minipage}
\caption{The figure shows experimental results for \textbf{intra-class drift} with (a) JakeFDB, (b) DebbieFDB, and (c) HarryFDB datasets. With all approaches, the accuracy for P6 increases as samples closer to P6 are included.}
\label{fig:intraclass}
\end{figure*}
\begin{figure*}[!h]
\centering
\begin{tabular}{cccc}
\includegraphics[width=0.31\textwidth]{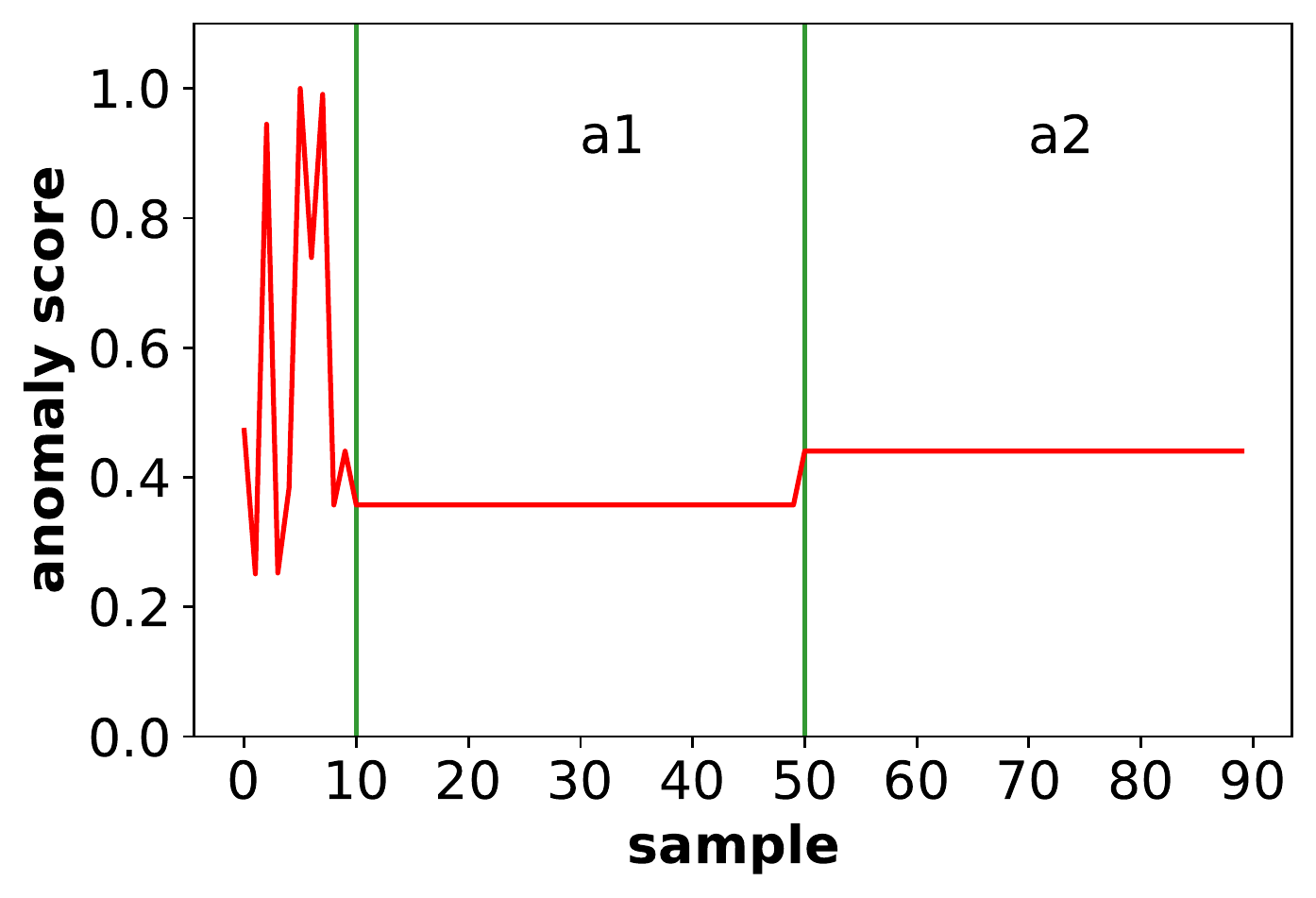} &
\includegraphics[width=0.31\textwidth]{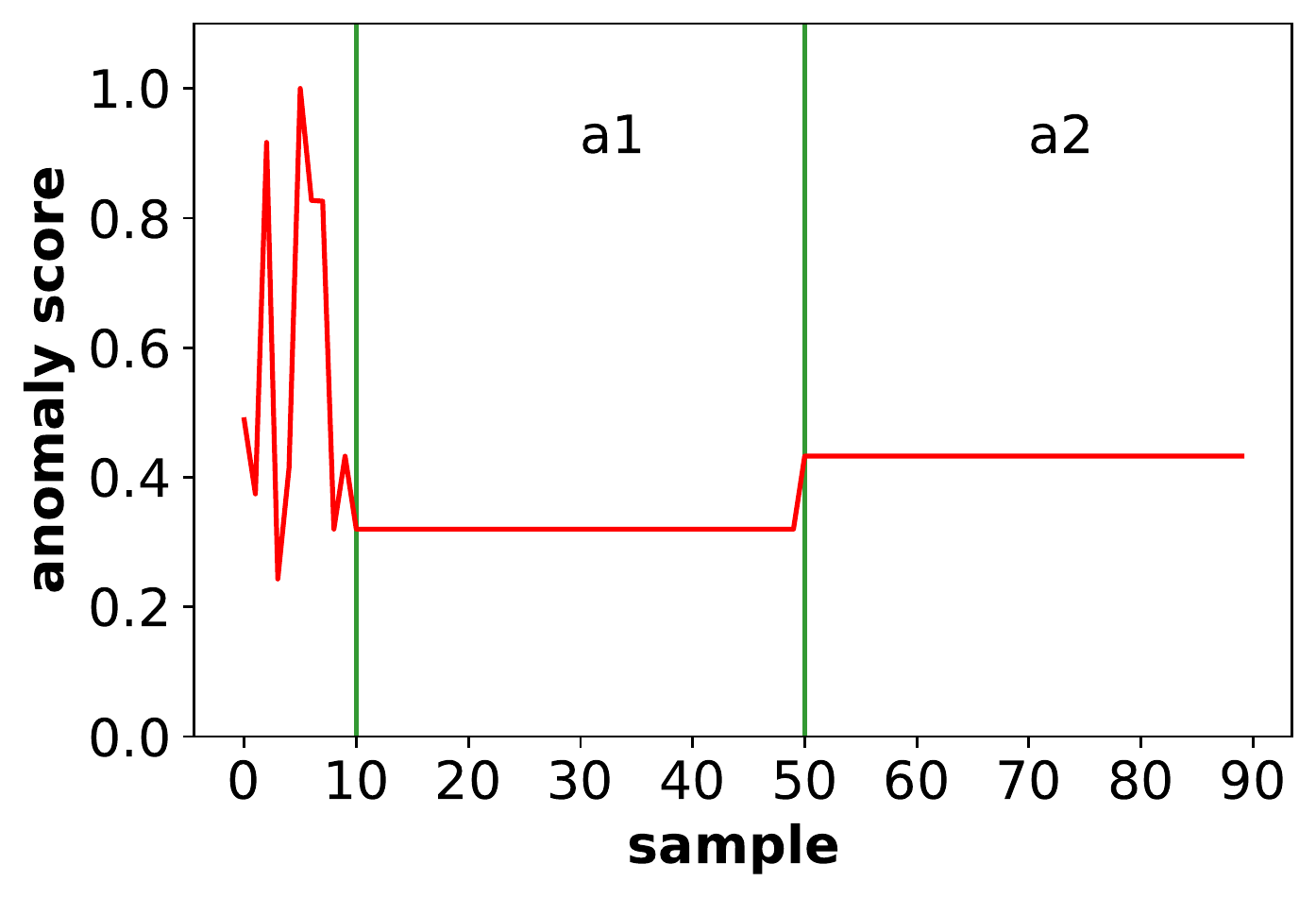} &
\includegraphics[width=0.31\textwidth]{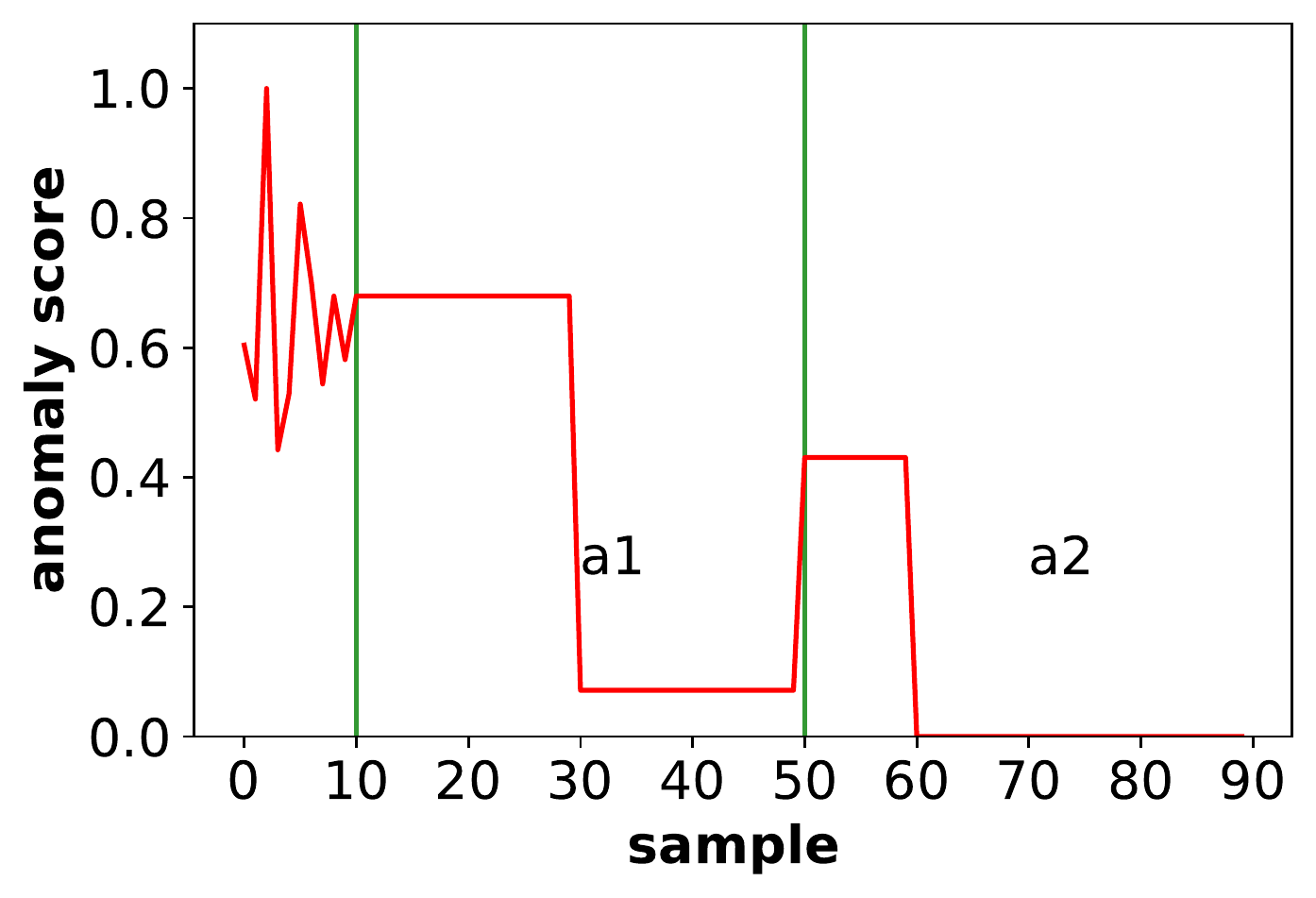} \\
(a) OCSVM  & (b) AE & (c) CMGMM  \\[6pt]
\end{tabular}
\begin{tabular}{cccc}
\includegraphics[width=0.31\textwidth]{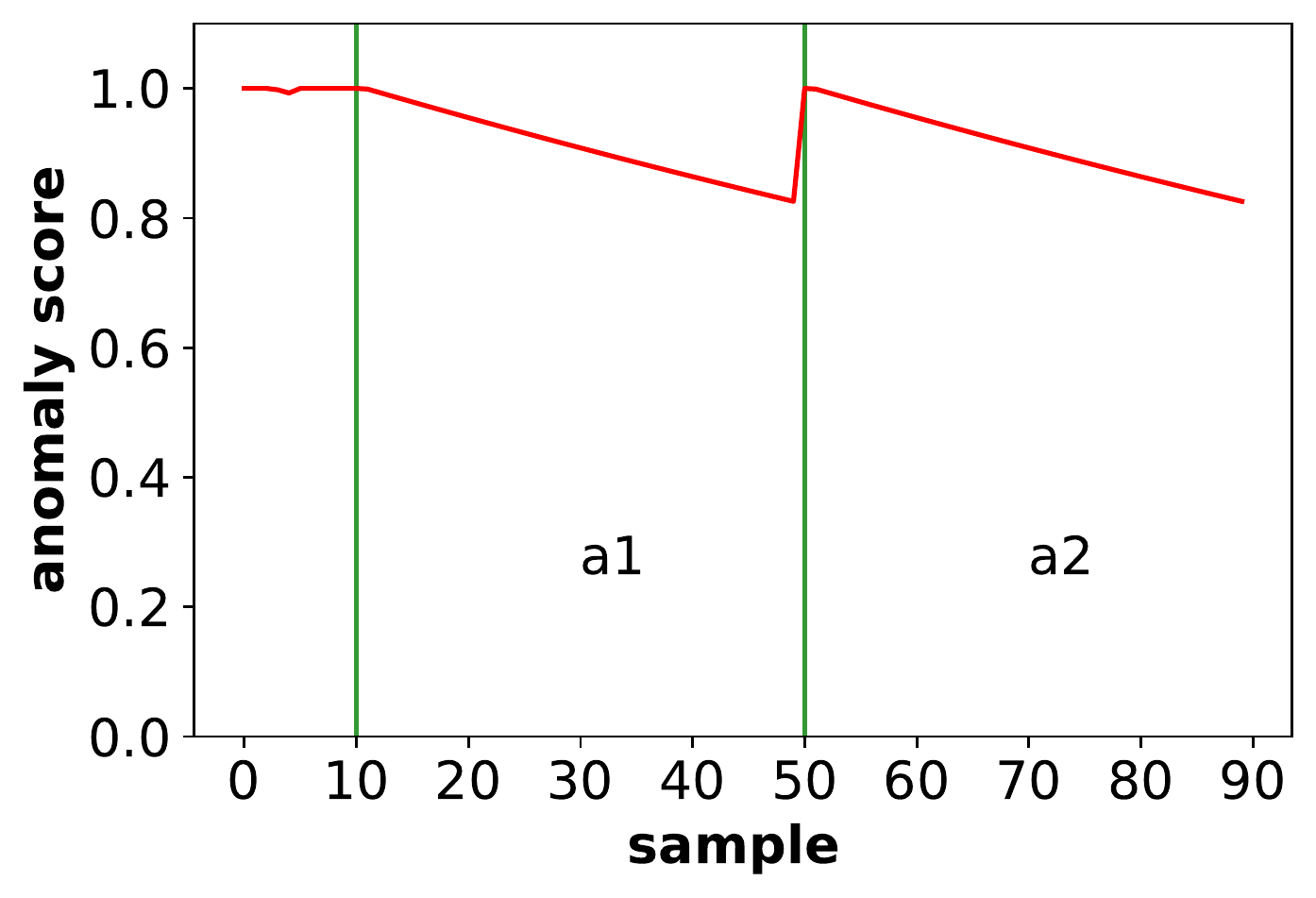} &
\includegraphics[width=0.31\textwidth]{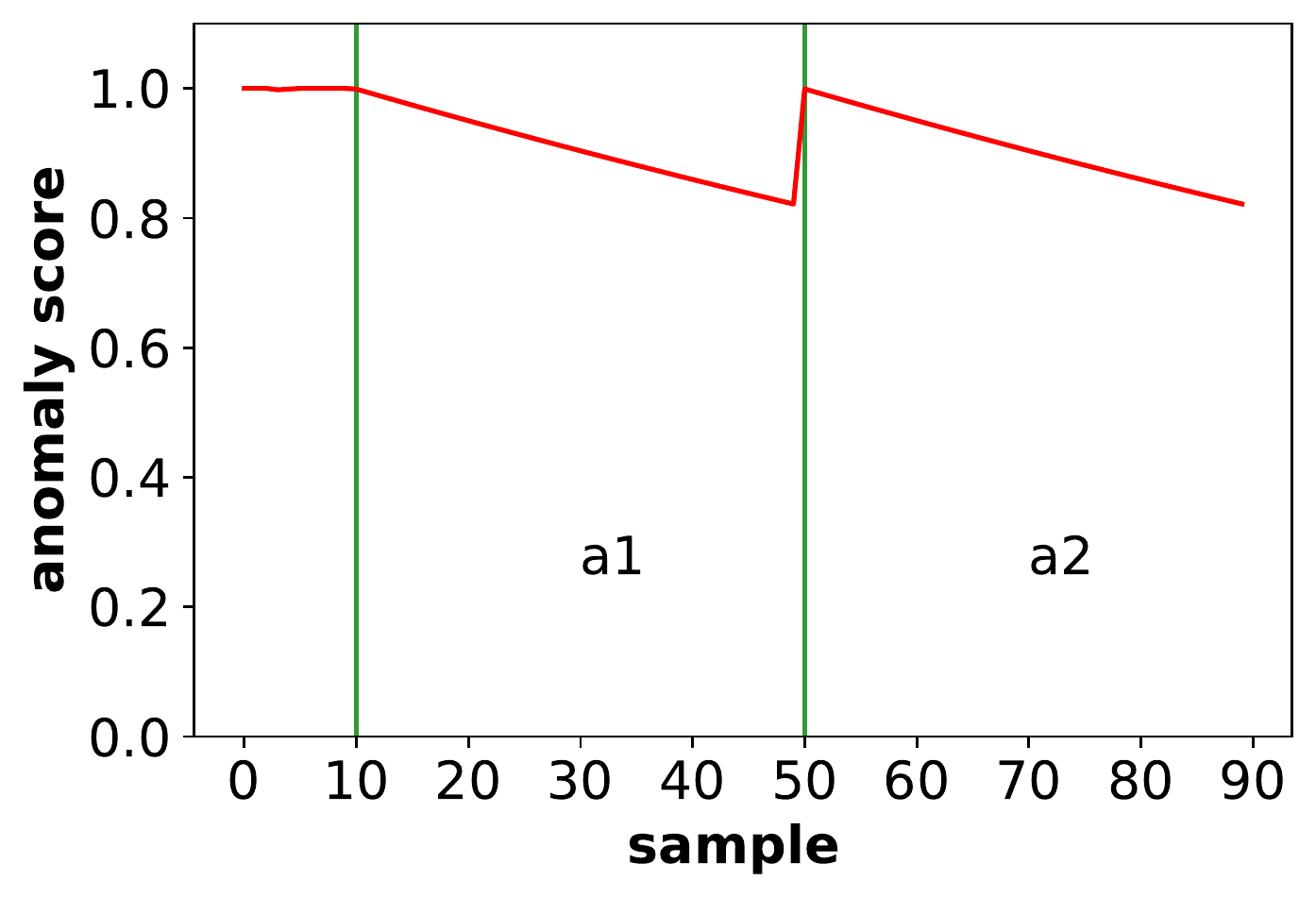} \\
(d) AGMM  &(e) UAGMM  \\[6pt]
\end{tabular}
\caption{ The figures show drift from abnormal to normal class, an \textbf{inter-class drift}, with different approaches.}
\label{fig:interclass}
\end{figure*}
\section{EXPERIMENTS \& RESULT ANALYSIS}\label{sec:exp}
In this section, we discuss the experimental setup, implementation detains and various experiments carried out to validate the efficacy of the proposed approach.
\subsection{Experimental setup and Implementation details}
For each dataset, we concatenate all the normal class samples from each season temporally and divide them into six equal parts denoted as P1, P2, P3, P4, P5, and P6. Thus each part shows incremental drift from the previous part. We get 606, 289, and 401 normal class samples in each part for JakeFDB, DebbieFDB, and HarryFDB, respectively. In the below sections, we refer to these parts while explaining the experiments and results.

The number of components in the PCA model is set to 20. For a feature size of 20, $\theta_{match}$ is set to 4.8, which corresponds to 68\% population~\cite{gallego2013mahalanobis}. For simplicity, we set fixed values of the majority of hyper-parameters used in the AGMM~\cite{kumari2020multivariate} and UAGMM approaches. These are as follows: $w_{0}$=0.001, $G$=0.95, and $Z$=4 times variance computed on P1. The above values are kept the same for all datasets and experiments involving AGMM or UAGMM. Further, in UAGMM, $\theta_{Bhat}$ is kept as 0.95 for all datasets.
\subsection{Intra-class drift} 
Any change in data distribution within a class is referred to as intra-class drift. 
Due to this drift in the normal class, it may appear that a normal sample from the late future does not appear to be a sample generated from the normal but the abnormal class; hence, it becomes important to handle this drift. To experimentally show the presence of drift in normal class and hence the importance of adaptive learning, the following experiments are designed:
\\(a) train on P1 (split no. 1) and test on P6
\\(b) train on P1+P2 (split no. 2) and test on P6
\\(c) train on P1+P2+P3 (split no. 3) and test on P6
\\(d) train on P1+P2+P3+P4 (split no. 4) and test on P6
\\(e) train on P1+P2+P3+P4+P5 (split no. 5) and test on P6.

If there is a gradual drift in the normal class data, the count of predicted normal samples from test data should increase owing to the growing training data over the subsequent splits.

We carry out the above experiments on static as well as adaptive learning models. 

In the static learning case, we consider One-class Support Vector Machine (OCSVM) and auto-encoders (AE) as they are primarily used for unsupervised anomaly detection.
The input to the OCSVM is a deep face descriptor extracted using the VGG-face model. The OCSVM model is trained and tested on deep features extracted using the VGG-face model. The trained OCSVM assigns a negative sign to the outliers, i.e., anomaly samples, and a positive sign to the normal samples. The baseline auto-encoder architecture is taken from the video anomaly detection approach by Wang et al.~\cite{wang2018detecting}. We fed the same deep features to this AE-based approach too. A higher reconstruction loss indicates the presence of anomalies. A threshold is then used to assign a binary class label as normal or abnormal.

In the case of adaptive learning, the CMGMM-based anomaly detection approach~\cite{id2020concept,id2022concept}, the AGMM-based anomaly detection approach~\cite{kumari2020multivariate} with $K$=5 and 10, and the proposed UAGMM are considered. In CMGMM, the hyper-parameters and other settings are kept the same as in ~\cite{id2020concept,id2022concept}. The number of components to train the best model using the EM algorithm ranged from 3 to 30. The data window is kept as 45. If the likelihood of a sample is high, then it is regarded as normal here. 
In AGMM based approach, the status of the Gaussian to which the sample is matched is considered as the model prediction. If the Gaussian to which the sample is matched is a normal class Gaussian, then it is a normal class sample or else an abnormal class sample. For AGMM and UAGMM, we keep $\alpha$ as 0.01, 0.001, and 0.0005 for JakeFDB, DebbieFDB, and HarryFDB datasets, respectively.

The accuracy is computed as the count of predicted samples into normal class divided by the maximum count achieved after P5 (split no. 5). Accuracy values achieved in all the experiments mentioned above is reported for JakeFDB, DebbieFDB, and HarryFDB in Figures~\ref{fig:intraclass} (a), (b), and (c), respectively.
We can see that for each approach, the accuracy shows a monotonically increasing trend. The aim here is not to show which approach works better but to show that performance improves as the model sees more and more samples close to test data and hence the necessity of adaptive learning. From these tables, it is evident that there is a presence of drift in normal data. If a one-time trained model is deployed for usage, its performance may be compromised on future data, as evident from the performance with experiment set (a). Frequent retraining, as done in the case of static learning-based approaches here, is not scalable with the growing training data. Therefore, adaptive learning should be followed. 
\subsection{Inter-class drift}
Here, we examine the behavior of static and adaptive learning-based approaches when there is inter-class drift. As explained in Section~\ref{sec:intro}, a sample may change its class from abnormal to normal and vice-versa, depending upon the frequency of its occurrence. We consider the case where an abnormal sample shows a drift to normal class, i.e., its anomaly score reduces over time. For this, we take P1 data from JakeFDB and train various static and adaptive models individually. During the testing phase, all the abnormal samples are fed. Further, the last two abnormal samples (let's name them a1 and a2) are repeated 40 times one by one. Note that during the test time, static models do not update their parameters, while the adaptive models update their parameters to accommodate the current data. 
In contrast to AGMM and UAGMM, where the parameters are updated at the arrival of each sample, in CMGMM, the parameter are updated only upon detecting drift in data windows. In this experiment, we keep the window size as 30 instead of 45 to show the changes in anomaly scores of a1 and a2.

\begin{figure*}[!h]
\begin{minipage}[b]{1\linewidth}
  \centering
 \includegraphics[width = 0.9\linewidth]{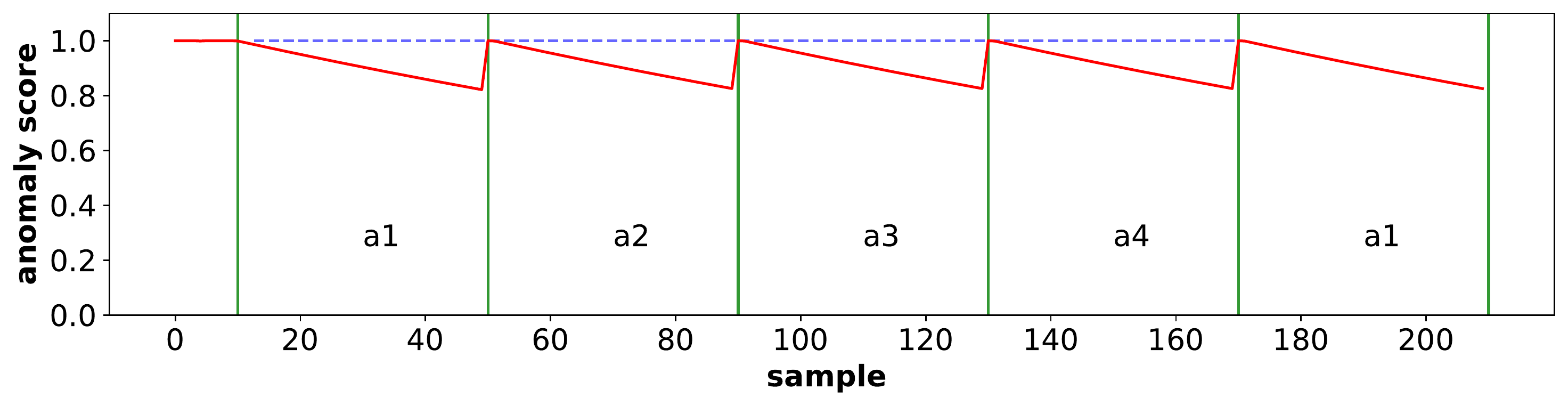}
   \centerline{(a) AGMM (K=4)}\medskip
\end{minipage}
\begin{minipage}[b]{1\linewidth}
  \centering
 \includegraphics[width = 0.9\linewidth]{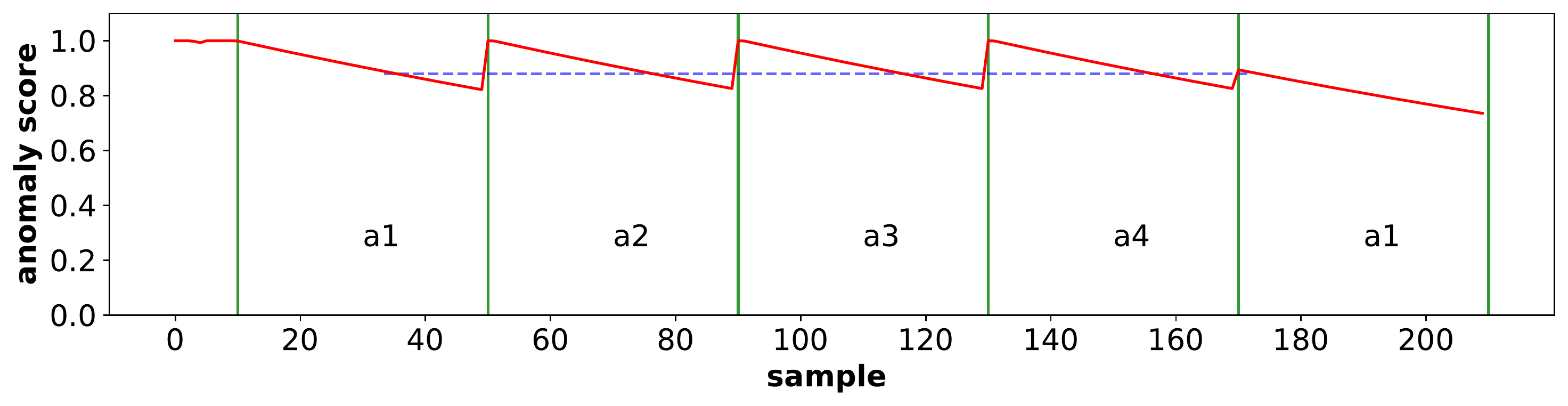}
   \centerline{(b) AGMM (K=5)}\medskip
\end{minipage}
\begin{minipage}[b]{1\linewidth}
  \centering
 \includegraphics[width = 0.9\linewidth]{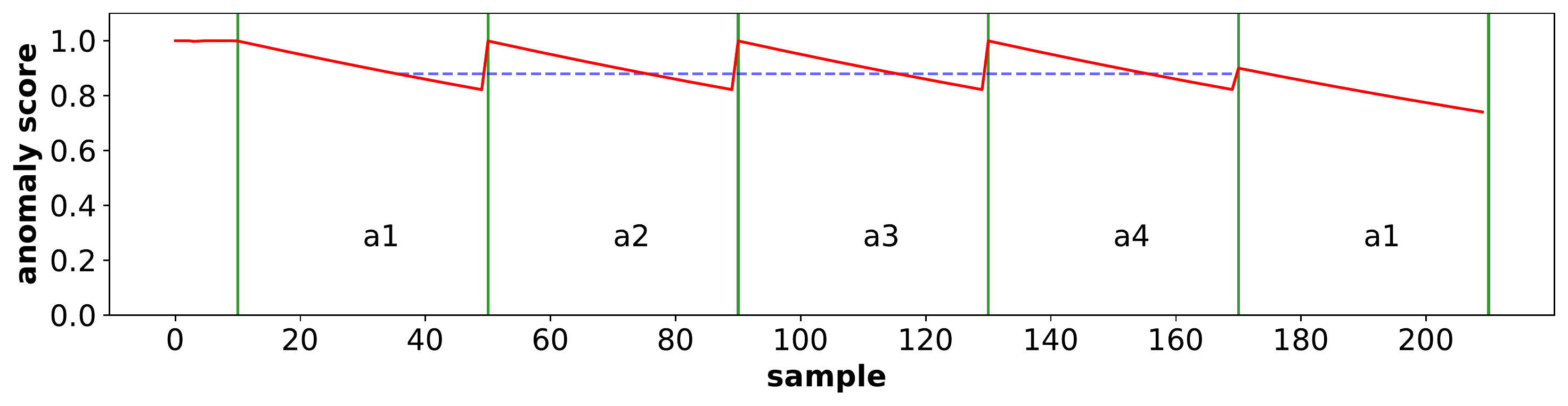}
   \centerline{(c) UAGMM}\medskip
\end{minipage}
\caption{\textbf{AGMM vs UAGMM:} deletion policy in AGMM cause loosing the inter-class transition for sample a1. Here, K=5 is suitable for AGMM.}
\label{fig:agmmVSuagmm_a4}
\end{figure*}
We show the anomaly scores on test samples given by these approaches in Figure~\ref{fig:interclass}. In the case of static learning-based models, i.e., OCSVM and AE, we see a constant anomaly score for all occurrences of a1 and a2. 
While, with adaptive approaches, i.e., CMGMM, AGMM, and UAGMM, the score decreases over time. However, with CMGMM, the score does not decrease smoothly as the model is updated at least after a window of samples is seen. Whereas with AGMM and UAGMM, the model is updated at the arrival of each sample. Thus, the anomaly scores slowly decrease as the model encounters more and more occurrences of the same sample, which in turn shows its transition from abnormal to normal class. We conclude that adaptive approaches can handle the inter-class drift, whereas a one-time trained model does not change its predicted score for a sample in the testing phase. Thus, a static model is not suitable in cases where there can be inter-class drift. Moreover, a model should consider modeling both normal as well as abnormal; otherwise, the drift from abnormal to normal can be missed.
\begin{figure*}[!h]
\begin{minipage}[b]{1\linewidth}
  \centering
 \includegraphics[width = 0.9\linewidth]{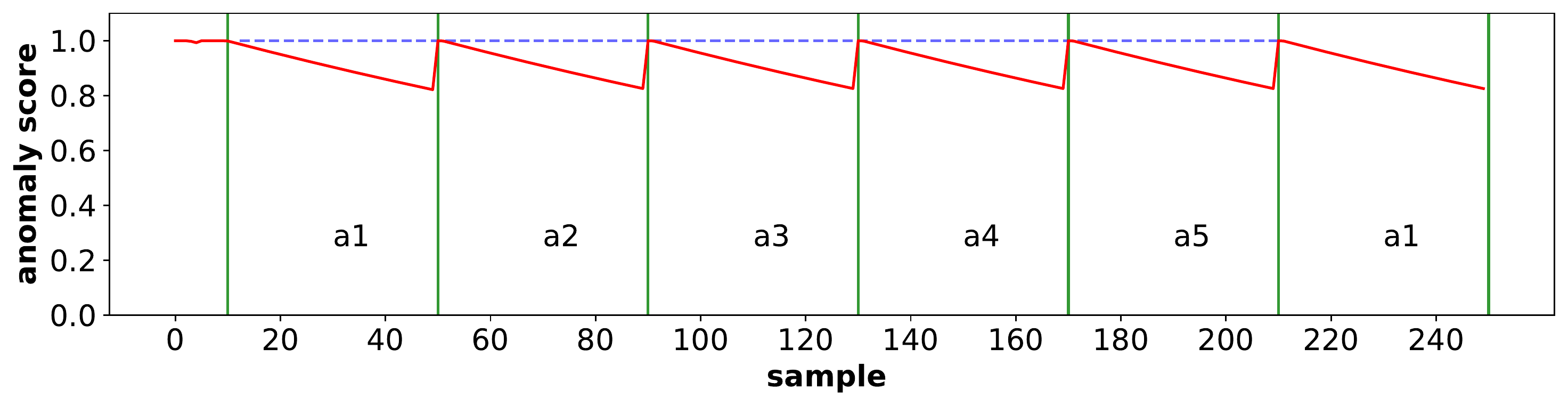}
   \centerline{(a) AGMM (K=5)}\medskip
\end{minipage}
\begin{minipage}[b]{1\linewidth}
  \centering
 \includegraphics[width = 0.9\linewidth]{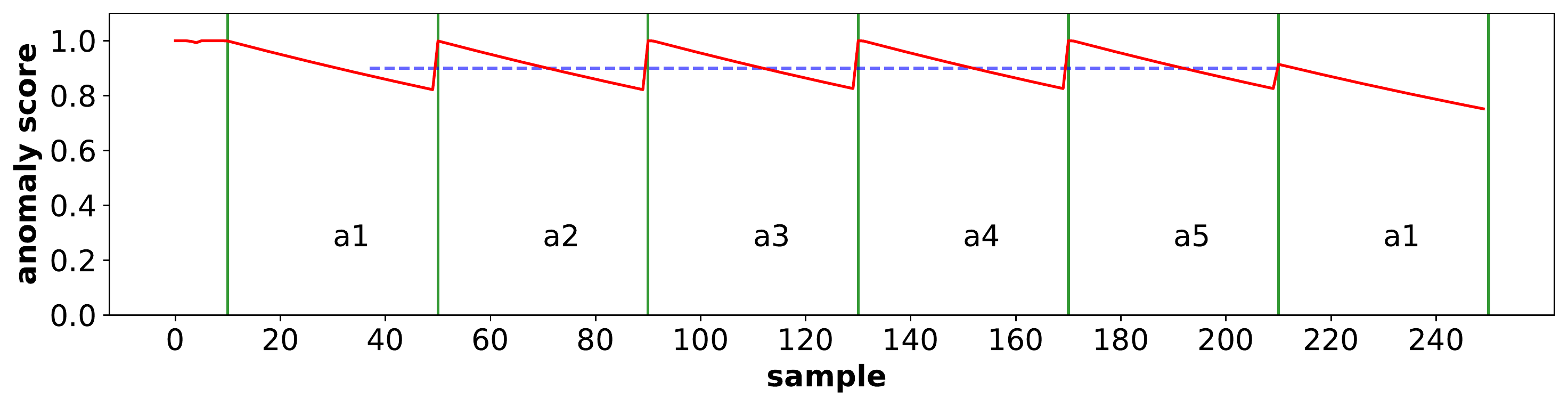}
   \centerline{(b) AGMM (K=6)}\medskip
\end{minipage}
\begin{minipage}[b]{1\linewidth}
  \centering
 \includegraphics[width = 0.9\linewidth]{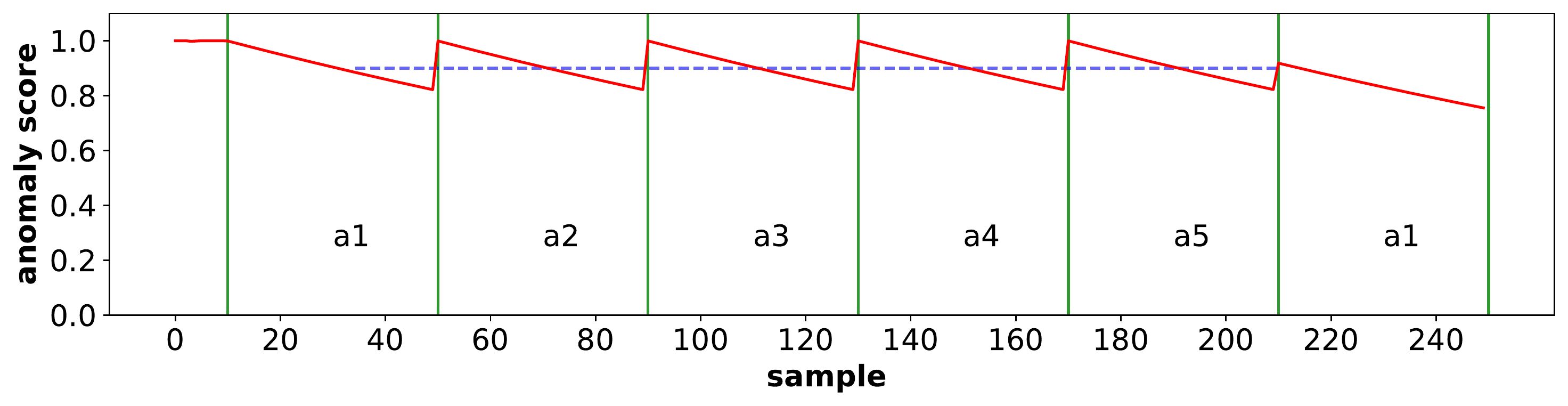}
   \centerline{(c) UAGMM}\medskip
\end{minipage}
\caption{\textbf{AGMM vs UAGMM:} deletion policy in AGMM cause loosing the inter-class transition for sample a1. Here, K=6 is suitable for AGMM.}
\label{fig:agmmVSuagmm_a5}
\end{figure*}
\begin{figure*}[!h]
\begin{minipage}[b]{1\linewidth}
  \centering
 \includegraphics[width =0.9\linewidth]{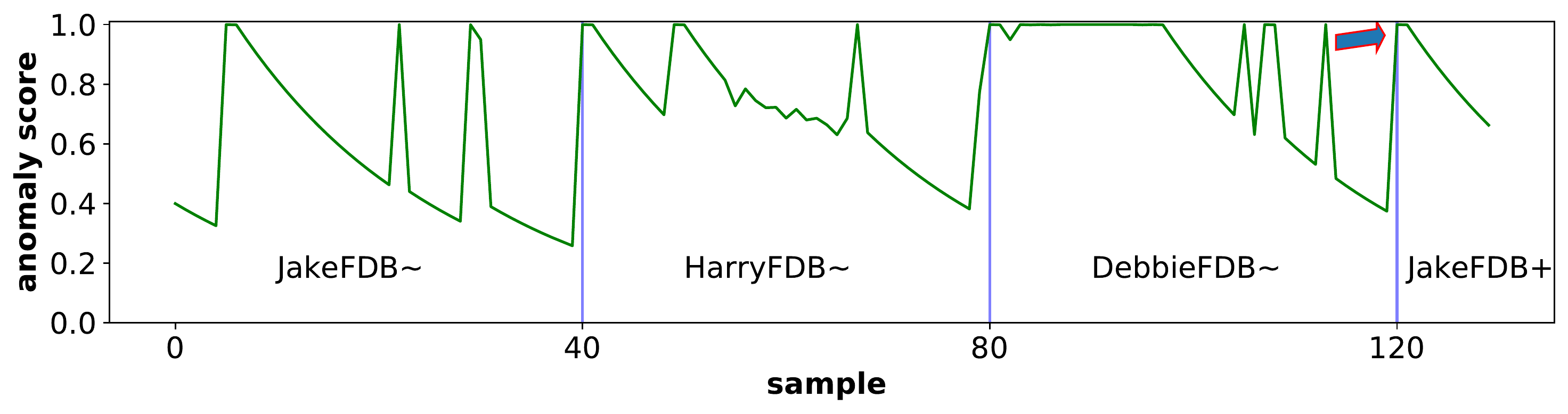}
   \centerline{(a) AGMM (K=4)}\medskip
\end{minipage}
\begin{minipage}[b]{1\linewidth}
  \centering
 \includegraphics[width =0.9\linewidth]{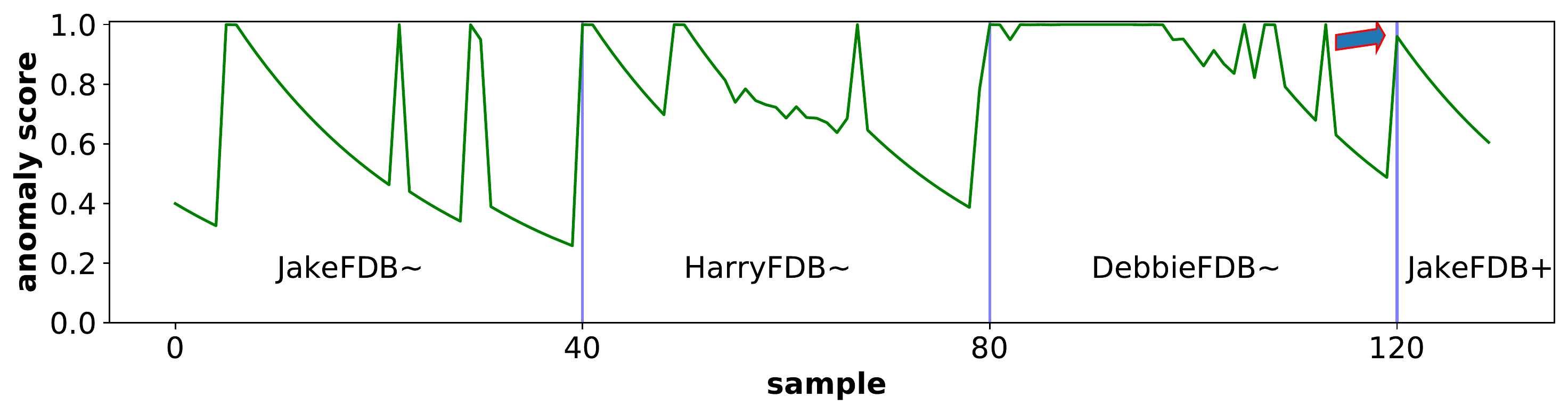}
   \centerline{(b) AGMM (K=6)}\medskip
\end{minipage}
\begin{minipage}[b]{1\linewidth}
  \centering
 \includegraphics[width = 0.9\linewidth]{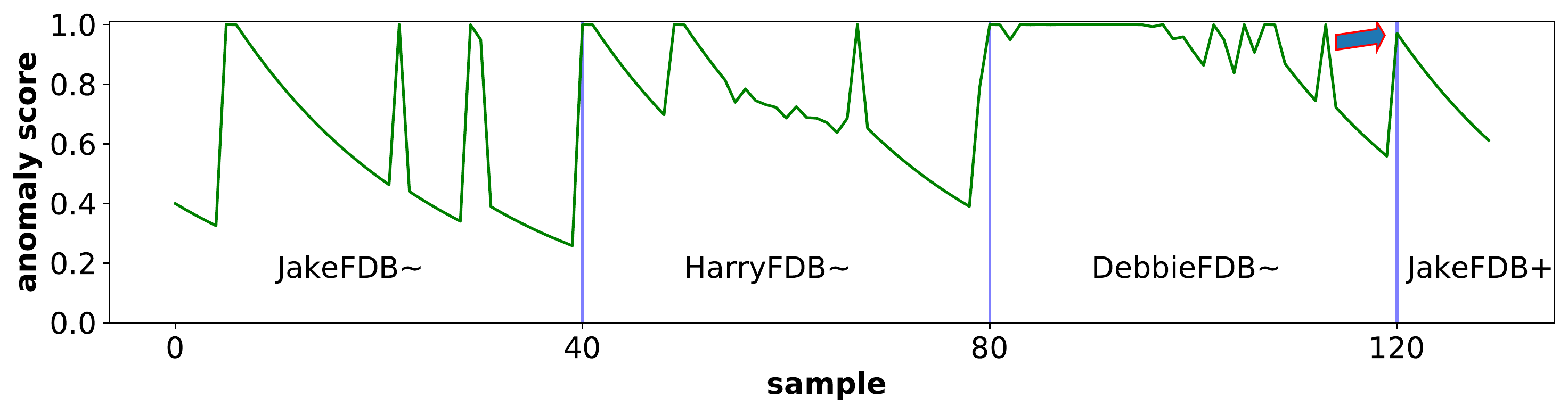}
   \centerline{(c) UAGMM }\medskip
\end{minipage}
\caption{\textbf{AGMM vs UAGMM:} Anomaly score for normal class samples. Samples from JakeFDB are shown after a long time, AGMM with $K=4$ has completely forgotten `Jake Harper' whereas in case of AGMM with $K=5$ or UAGMM the corresponding mode is present in the mixture.}
\label{fig:agmmVSuagmm_normalShift}
\end{figure*}
\subsection{AGMM vs. UAGMM}
We have seen that both the adaptive approaches, i.e., AGMM~\cite{kumari2020multivariate} and UAGMM are suitable for anomaly detection in cases inter-class or intra-class drift is present. However, there is a challenge with the existing baseline AGMM-based anomaly detection approach. If a new mode is to be created to accommodate a new anomaly sample and the number of Gaussians has reached the maximum limit (K), the new Gaussian is added by deleting the weakest Gaussian from the mixture. This can cause frequent deletion of modes in a dynamic environment which can lead to loss of transition from abnormal to normal for a given sample.

We experimentally show the disadvantage of the deletion policy through Figures~\ref{fig:agmmVSuagmm_a4} and~\ref{fig:agmmVSuagmm_a5}. We fed P1 data of JakeFDB followed by all the abnormal samples to both the models. Then, the last four abnormal samples, namely a1, a2, a3, and a4, are repeated one by one, each one 40 times. Then, a1 is again repeated 40 times. We show anomaly scores of these abnormal samples given by both adaptive approaches in Figure~\ref{fig:agmmVSuagmm_a4}.

For AGMM with $K$=4 (Figure~\ref{fig:agmmVSuagmm_a4} (a)), we see that when a1 is repeated after a4, it gets a score of 1. With $K$=4, modes for normal class, a2, a3, and a4 are present, and the mode corresponding to a1 has been deleted in spite of its frequent occurrence in the recent past. 
If we increase the limit on maximum possible Gaussians from 4 to 5, as shown in Figure~\ref{fig:agmmVSuagmm_a4} (b), we observe that a1 gets a score lower than 1, which indicates that the mode for a1 has not been deleted. Thus, the transition for sample a1 is not discarded. However, experimentally setting a suitable number of $K$ is not scalable as the number of distinct object/ event present at any time are unknown and dynamic. If instead of repeating just a1 to a4, when we repeat a1 to a5, the least number of modes required, which will not cause deletion of a1, changes. As shown in Figure~\ref{fig:agmmVSuagmm_a5} (a), with $K$=5 too, the score for a1 at its occurrence after a5 is 1. But when we increase $K$ from 5 to 6, the mode for a1 is not deleted, and it gets a score lower than 1.

On the other hand, with UAGMM, whether we fed a1 to a4 or a1 to a5, it does not cause the deletion of modes permanently. It forgets only in the weight normalization step but with a slow pace. It can be seen from Figure~\ref{fig:agmmVSuagmm_a4} (c) and Figure~\ref{fig:agmmVSuagmm_a5} (c) that sample a1 gets a score lower than 1 at its occurrence after a4 as well as a5. Thus, with UAGMM explicitly setting a suitable $K$ from time to time is not required.

Further, another set of experiments with only normal data is carried out. We took 40 initial samples from each of the three datasets. We name them as $\sim$JakeFDB, $\sim$HarryFDB, and $\sim$DebbieFDB. Further, we take initial 10 samples from JakeFDB and name as $\sim$JakeFDB+. They are fed in the same order to each AGMM and UAGMM. The predicted anomaly scores for AGMM with $K$=4, AGMM with $K$=6, and UAGMM are shown in Figures~\ref{fig:agmmVSuagmm_normalShift} (a), (b), and (c), respectively. We can see that both AGMM and UAGMM show a higher score for each dataset initially followed by a decreasing trend. Some spikes or non-decreasing trend in between is also present. Since the samples are not collected in controlled and ideal settings, some samples may appear as false anomalies. Moreover, the samples may not be correctly aligned on the temporal axis as the shooting of episodes may not be done in order. The challenges associated with these datasets are also mentioned in Table~\ref{tab:realVSIdeal}.

However, these shortcomings do not affect this experiment. From Figures~\ref{fig:agmmVSuagmm_normalShift} (b) and (c), we  observe that when samples of ‘Jake Harper’ are shown to the model after a long time, i.e., after the samples of ‘Harry Potter' and ‘Debbie Gallagher’, a score lower than 1 is predicted. 
However, in the case of AGMM with a lower $K$ (see Figure~\ref{fig:agmmVSuagmm_normalShift} (a)), ‘Jake Harper’ gets 
a score of 1 and hence an anomaly. It means the mode for ‘Jake Harper’ has been removed from the mixture due to its deletion policy. For AGMM with a larger $K$, the mode is not yet deleted but susceptible to permanent deletion in the near future. However, with UAGMM, the mode is never completely removed from the mixture but slowly forgotten. Therefore,
we can say that UAGMM reassembles more with how the human brain retains information for longer. Current information is given more weightage whereas older ones are given less weightage. Thus we conclude that UAGMM is more suitable for anomaly detection than AGMM for concept drift cases.
\begin{table*}[!h]
\centering
\caption{Ideal vs. real case for facial dataset collection}
 \label{tab:realVSIdeal}
\begin{tabular}{|c|c|}
\hline
 Ideal case & Real/ practical case \\
 \hline
 
\begin{tabular}[c]{@{}c@{}}Face images are captured on\\ approximately equal time interval \end{tabular}&\begin{tabular}[c]{@{}c@{}}Not taken on approximately equal time interval.  \\ As some episode or season are shooted with less time-gap, whereas some are \\shooted after a longer time-gap (sometimes upto more than a year too).\\Some episodes might have not been shooted in a temporal sequence.  \end{tabular}  \\\hline

\begin{tabular}[c]{@{}c@{}}Unaltered faces, i.e., faces are mostly \\similar in two or more subsequent capture\end{tabular}&\begin{tabular}[c]{@{}c@{}}At subsequent appearances, different type of \\heavy make-ups, lighting conditions, expression, etc., might present. \end{tabular} \\\hline

Similar imaging set-up& At subsequent appearances, the camera might be at different angle, distance, etc. \\\hline

\end{tabular}
\end{table*}
\begin{figure*}[!h]
\centering
\includegraphics[scale=0.65]{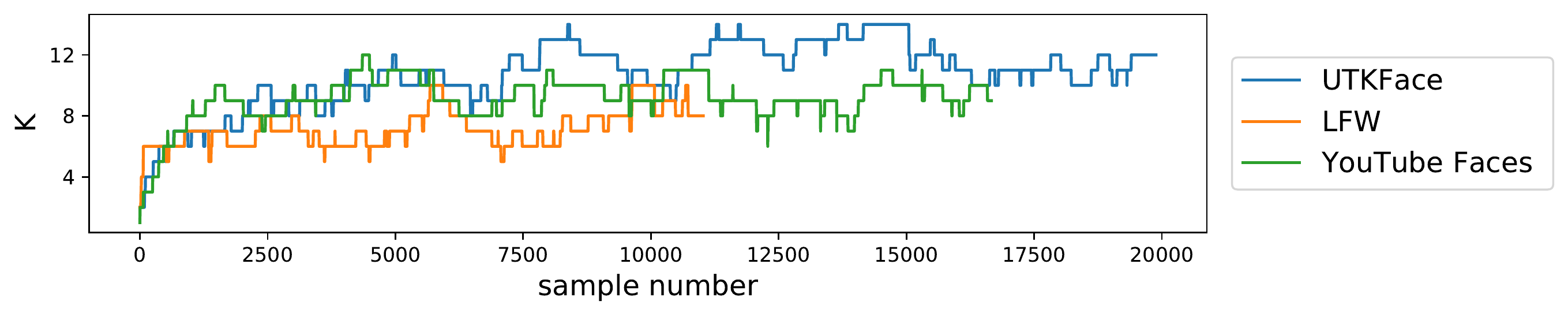}
\caption{Value of $K$ over the time for UTKFace~\cite{zhang2017age}, LFW~\cite{huang2012learning}, and YouTube Faces~\cite{wolf2011face} datasets}
\label{fig:noOfmodes}
\end{figure*}
\subsection{Unconstrained memory}
Contrary to AGMM, we do not need to know the suitable number of maximum allowed Gaussians, i.e., $K$, with UAGMM based framework. A suitable value of $K$ can be predetermined by analyzing some past data; however, the scene/object/situation under surveillance might change in the future, and the pre-defined value of $K$ might not be suitable. Thus, it can lead to degradation in performance as some contexts may be deleted. Instead, an adaptive value of $K$ as in our framework will be more suitable for the life-long application of the model in a dynamic environment. In the proposed framework, we do not need to specify $K$; it shrinks or grows as per the current dynamics. We experimentally show that number of Gaussians does not explode even if we keep it in an unconstrained manner. We considered three widely used public facial datasets, viz., UTKFace~\cite{zhang2017age}, LFW~\cite{huang2012learning}, and YouTube Faces~\cite{wolf2011face}, for this experiment. We pass these datasets in our framework and report the value of $K$ over the samples in Figure~\ref{fig:noOfmodes}. For each dataset, the same hyper-parameters are taken. We can see that for all the datasets, the value of $K$ increases initially, followed by a convergence trend. We can see some spikes later on, but the value of $K$ shrinks and does not lead to an explosion due to the merge facility. Thus keeping no limit on the maximum allowed number of Gaussians does not create an overflow in terms of memory.
\section{Conclusion}\label{sec:conc}
In this paper, we described a novel anomaly detection framework and datasets with the objective of detecting anomalies from streaming multimedia data having concept drift. For anomaly detection in the presence of concept drift, mainly the baseline AGMM that facilitates adaptive learning is utilized~\cite{kumari2020multivariate}. However, the constrained memory in baseline AGMM causes frequent deletion of modes, which in turn can cause missing the inter-class transitions. The proposed UAGMM-based approach proved to handle the concept drift in data better compared to the baseline AGMM. Experimentally, we demonstrated that the AGMM performs inadequately when the model memory limit is smaller than the number of recent unique samples or when the dataset contains events with a longer temporal context. 
We conclude that the proposed UAGMM can be employed in place of AGMM for anomaly detection in situations involving concept drift.
\bibliographystyle{IEEEtran}
\bibliography{main.bib}
\end{document}